%% file: 0-main.tex
\documentclass[10pt,twocolumn,letterpaper]{article}

\usepackage[pagenumbers]{cvpr} 

\usepackage{graphicx}
\usepackage{amsmath}
\usepackage{amssymb}
\usepackage{booktabs}
\usepackage{array}
\usepackage{multirow}
\usepackage{nicefrac}       
\usepackage{microtype}  
\usepackage{tabularx}
\usepackage[toc,page]{appendix}
\usepackage[dvipsnames]{xcolor}
\usepackage[accsupp]{axessibility} 
\usepackage[pagebackref,breaklinks,colorlinks]{hyperref}

\usepackage[capitalize]{cleveref}
\crefname{section}{Sec.}{Secs.}
\Crefname{section}{Section}{Sections}
\Crefname{table}{Table}{Tables}
\crefname{table}{Tab.}{Tabs.}

\newcommand{\ours}[0]{DIVeR}
\newcommand{\textbl}[1]{\emph{\underline{#1}}}

\def\cvprPaperID{} 
\def\confName{CVPR}
\def\confYear{2022}

\usepackage{capt-of,etoolbox}	
\makeatletter
\apptocmd\@maketitle{{\myfigure{}\par}}{}{}
\makeatother

\begin{document}
\newcommand\myfigure{%
\centering
\vspace*{-0.15in}
}
\title{DIVeR: Real-time and Accurate Neural Radiance Fields \\ with Deterministic Integration for Volume Rendering}

\author{
Liwen Wu
\and 
Jae Yong Lee
\and 
Anand Bhattad  
\and 
Yu-Xiong Wang 
\and 
David Forsyth
\and
University of Illinois at Urbana-Champaign\\
{\tt\small \{liwenwu2, lee896, bhattad2, yxw, daf\}@illinois.edu}
}
\maketitle

\input{1-abs}
\input{2-intro}
\input{3-related_work}
\input{figures/pipeline}
\input{4-method}
\input{5-experiments}
\input{6-conclusion}

\paragraph{Acknowledgements:}
We thank Sara Aghajanzadeh, Derek Hoiem, Shenlong Wang, and Zhen Zhu for their detailed and valuable comments on our paper.

{\small
\bibliographystyle{ieee_fullname}
\bibliography{references.bib}
}

\newpage
\begin{appendix}
\input{appendix/implementation_details}
\input{appendix/experiment_details}
\end{appendix}

\end{document}

%% file: 1-abs.tex
\begin{abstract}
DIVeR builds on the key ideas of NeRF and its variants -- density models and volume rendering -- to learn 3D object models that can be rendered realistically from small numbers of images.  In contrast to all previous NeRF methods, DIVeR uses deterministic rather than stochastic estimates of the volume rendering integral.  DIVeR's representation is a voxel based field of features.  To compute the volume rendering integral, a ray is broken into intervals, one per voxel; components of the volume rendering integral are estimated from the features for each interval using an MLP, and the components are aggregated.  As a result, DIVeR can render thin translucent structures that are missed by other integrators.  Furthermore, DIVeR's representation has semantics that is relatively exposed compared to other such methods -- moving feature vectors around in the voxel space results in natural edits. Extensive qualitative and quantitative comparisons to current state-of-the-art methods show that DIVeR produces models that (1) render at or above state-of-the-art quality, (2) are very small without being baked, (3) render very fast without being baked, and (4) can be edited in natural ways. 
Our real-time code is available at: \url{https://github.com/lwwu2/diver-rt}
\end{abstract}

%% file: 2-intro.tex
\section{Introduction}
\label{sec:intro}
Turning a small set of images into a renderable model of a scene is an important step
in scene generation, appearance modeling, relighting, and computational photography.
The task is well-established and widely studied; what form the model should take is
still very much open, with models ranging from explicit representations of geometry and
material through plenoptic function models~\cite{Adelson1991ThePF}.
Plenoptic functions are hard to smooth,
but neural radiance field (NeRF)~\cite{mildenhall2020nerf} demonstrates that a Multi Layer Perceptron (MLP) with positional encoding
is an exceptionally good smoother, resulting in an explosion of variants (details in related work).
All use one key trick: the scene is modeled as density and color functions, rendered using stochastic estimates of
volume rendering integrals. We describe an alternative approach, \emph{deterministic integration for volume rendering}
(\ours), which is competitive in speed and accuracy with the state of the art. 

\input{figures/nerf_fail}

We use a deterministic integrator because stochastic estimates
of integrals present problems. Samples may miss important effects (Fig.~\ref{fig:nerf-fail}). Fixing this by increasing the sampling rate is costly: accuracy improves slowly in the number of samples $N$,
(for Monte Carlo methods, standard deviation goes as $1/\sqrt{N}$~\cite{boyle}), but the cost of computation grows linearly.
In contrast, our integrator
combines per-voxel estimates of the volume rendering integral into a
single estimate using alpha blending (Sec.~\ref{subsec:dvrf}).

Like NSVF~\cite{liu2020neural}, we use a voxel based representation of the color and density. Rather than represent
functions, we provide a feature vector at each voxel vertex. The feature vectors at the vertices of a given voxel are
used by an MLP to compute the deterministic integrator estimate for the section of any ray passing through the voxel. Mainly, a model is learned by gradient descent on feature vectors and MLP parameters to minimize the prediction error
for the training views; it is rendered by querying the resulting structure with new rays. Sec.~\ref{sec:method} provides
the details. 

Given similar computational resources, our model is efficient to train, likely because the deterministic integration can
fit the integral better, and there is no gradient noise produced by stochastic integral estimates. As
Sec.~\ref{sec:experiments} shows, the procedure results in very small models ($\sim 64 \text{MB}$) which render very
fast ($\sim 50$ FPS on a single 1080 Ti GPU) and have comparable PSNR with the best NeRF models. 

%% file: figures/nerf_fail.tex
\begin{figure}[t]
    \centering
    \setlength\tabcolsep{1.5pt}
    \begin{tabular}{cc}
         \includegraphics[width=0.20\textwidth]{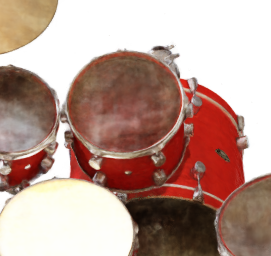}&
         \includegraphics[width=0.20\textwidth]{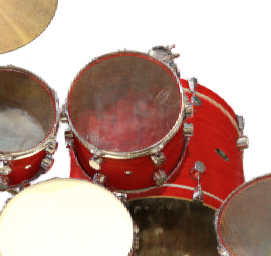}\\
         NeRF & PlenOctrees \\
         \includegraphics[width=0.20\textwidth]{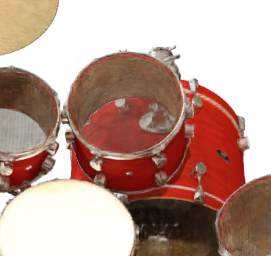}&
         \includegraphics[width=0.20\textwidth]{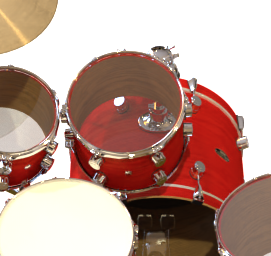}\\
         $\text{\ours}$  (Ours) &Ground Truth\\
    \end{tabular}
    \caption{\textbf{Monte Carlo vs. feature integration}. 
    Methods like NeRF \cite{mildenhall2020nerf} and
      PlenOctrees \cite{yu2021plenoctrees} that use Monte Carlo (stochastic) integrators
      fail to render the translucent drumhead;
      it is thin, and so is hit by few samples. NeRF's estimator will not model it with practical numbers of samples.
      In contrast, our method uses a deterministic integrator which
      directly estimates the section of volume rendering integral near
      the drum plane through feature integration
      (Sec.~\ref{subsec:dvrf}); this uses fewer calls to the integrator and
      still successfully models the transparency.
      }  
    \label{fig:nerf-fail}
\end{figure}

%% file: 3-related_work.tex
\section{Background}
\label{sec:background}
NeRF \cite{mildenhall2020nerf} represents 3D scenes with a density field $\sigma(\mathbf{x})$ and a color field $\mathbf{c}(\mathbf{x},\mathbf{d})$  which are
functions of 3D position $\mathbf{x}$ and view direction $\mathbf{d}$ encoded by
an MLP with weights $\mathbf{w}$. To render a pixel, a ray $\mathbf{r}(t)=\mathbf{o}+\mathbf{d}t$ is shot from the camera center $\mathbf{o}$ through the pixel center in direction $\mathbf{d}$ and follows the volume rendering equation \cite{10.1145/964965.808594} to accumulate the radiance:
\input{equations/vre_origin}
Closed-form solutions for Eq.~\ref{eq:vre-ori} are not available, so NeRF uses Monte Carlo integration by randomly sampling $n$ points $\mathbf{x}_i=\mathbf{r}(t_i),i=1,\ldots,n$ along the ray from eye to far with their radiance and density values $(\mathbf{c}_i,\sigma_i)=\text{MLP}_\mathbf{w}(\mathbf{x}_i, \mathbf{d}_i)$. The radiance and density function is then treated as constant in every interval $[t_{i-1}, t_{i}]$, and an approximation of the volume rendering equation is given as:
\input{equations/vre_nerf}
where $\alpha_i$ denotes the accumulated alpha values along the interval
and $\delta_i=\lVert \mathbf{x}_{i+1}-\mathbf{x}_{i}\rVert_2$ is the interval length. During training, NeRF learns to adjust the density and color fields to produce training images, which is achieved by optimizing weights $\mathbf{w}$ in respect to squared error between rendered pixel and its ground truth:
\input{equations/mean_squared_error}

\section{Related Work}
\label{sec:related}

\paragraph{Novel view synthesis:}
All scene modeling methods attempt to exploit regularities in the plenoptic function (the spectral radiance traveling
in any direction at any point) of a scene.  One approach is to compute an explicit geometric representations (point clouds
\cite{aliev2020neural,9064947}; meshes \cite{Riegler2020FVS,Riegler2021SVS,10.1145/3306346.3323035}) usually obtained
from some 3D reconstruction algorithms (\eg COLMAP \cite{schoenberger2016sfm}). The geometry can then carry
deep features, which are projected and then processed by neural networks to get the
image. Building an end-to-end optimizable pipeline is hard, however.  Alternatively, one could
use voxel grids\cite{Lombardi:2019,rematasCVPR20,sitzmann2019deepvoxels}, where scene observations are encoded as 3D features and
processed by 3D, 2D Convolutional Neural Networks (CNNs) to get the rendered images, yielding cleaner training but a very memory intensive model
which is unsuitable for high resolution rendering.

Multi-plane images (MPIs) \cite{Zhou2018StereoM,Srinivasan2019PushingTB,Mildenhall2019LocalLF,Wizadwongsa2021NeX} offer
novel views without requiring a precise geometry proxy.  One represents a scene as a set of parallel RGBA images and
synthesizes a novel view by warping the images to the target view and then alpha blending them; this fails when the view
changes are too large. Image based rendering (IBR)
approaches \cite{Wang2021IBRNetLM,chen2021mvsnerf,SRF} render a view by interpolating the nearby observations
directly. Most IBR methods generalize well to unseen data, such that a new scene in an IBR model can be rendered
off-the-shelf or with a few epochs of fine-tuning. 

An alternative is to represent proxies for the plenoptic function using neural networks, and then
raycast. \cite{sitzmann2019srns,yariv2020multiview,Kellnhofer:2021:nlr} use signed distance field like functions, and
\cite{Saito2019PIFuPI,Niemeyer2020DifferentiableVR} represent the scene geometry as an occupancy field. NeRF
\cite{mildenhall2020nerf} models the plenoptic function using an MLP to encode a density field (of position) and a color field (of
position and direction).  The radiance at a point in a direction is given by a volume rendering integral \cite{10.1145/964965.808594}. Training is done by adjusting the MLP parameters to produce the right answer for a given set of images.
The method  can produce photo-realistic rendering on complex scenes, including rendering transparent surfaces
and view dependent effects, but takes a long time to train and evaluate. 

NeRF has resulted in a rich collection of variants. \cite{boss2021nerd,nerv2021,Zhang2021NeRFactorNF} modify the NeRF to allow control of surface materials and lighting;
NeRF-W~\cite{martinbrualla2020nerfw} augments the inputs with image features to help resolve ambiguity between photos in
the wild.~\cite{park2021nerfies,pumarola2020d,park2021hypernerf} show how to model deformation, and~\cite{li2020neural,xian2021space,du2021nerflow,Gao-freeviewvideo,li2021neural} apply NeRF to 4D videos. Finally,~\cite{Wang2021IBRNetLM,SRF,chen2021mvsnerf,yu2020pixelnerf,tancik2020meta} try to improve the generalizability and
training speed, and~\cite{Chan2021piGANPI,Schwarz2020NEURIPS,grf2020,rematasICML21,Niemeyer2020GIRAFFE} adopt the
architecture to generative models.  

\vspace{-4mm}
\paragraph{Rendering NeRF faster:}
NeRF's stochastic integrator not only misses thin structures (which are hard to find with samples, Figure~\ref{fig:nerf-fail}), but
also presents efficiency problems.
 The main strategy for improving the efficiency of NeRF (as in any MC integrator) is coming up with better importance functions~\cite{boyle}. NSVF~\cite{liu2020neural}  significantly reduces the number of samples (equivalently, MLP calls; render time) by imposing a
voxel grid on the density, and then pruning voxels with empty density at training time. An alternative is a depth oracle that ensures that MLP samples occur
only close to points with high density~\cite{neff2021donerf}. AutoInt~\cite{autoint2021} further offers a more efficient estimator for the volume
rendering integral by constructing an approximate antiderivative (though the absorption integral must still be approximated), which allows
fewer MLP queries at rendering time. In contrast to these methods, we use a deterministic integral estimator.

But pure importance based methods cannot render in real-time, because they rely on MLPs that are relatively expensive to evaluate.
FastNeRF~\cite{garbin2021fastnerf} discretizes continuous fields into bins and caches the bins that have been evaluated
for subsequent frames.  PlenOctrees~\cite{yu2021plenoctrees} and SNeRG~\cite{hedman2021baking} pre-bake the
results of the NeRF into sparse voxels and use efficient ray marching to achieve interactive frame rate.
These methods achieve real-time rendering at a cost of noticeable loss in quality (ours does not), or of requiring a high-resolution voxel grid and so a large storage cost (ours does not).
Alternative strategies include: caching MLP calls into MPIs (Nex~\cite{Wizadwongsa2021NeX});
and speeding up MLP evaluation by breaking one MLP into many small local specialist MLPs (KiloNeRF~\cite{reiser2021kilonerf}).  In contrast, we use the representation and MLP obtained at training time.

%% file: equations/vre_origin.tex
\begin{equation}
    \hat{\mathbf{c}}(\mathbf{r})=\int_0^{\infty}e^{-\int_0^t \sigma(\mathbf{r}(\tau))d\tau}\sigma(\mathbf{r}(t))\mathbf{c}(\mathbf{r}(t),\mathbf{d})dt.
    \label{eq:vre-ori}
\end{equation}

%% file: equations/vre_nerf.tex
\begin{gather}
    \hat{\mathbf{c}}(\mathbf{r}) = \sum_{i=1}^{n} \prod_{j=1}^{i-1} (1-\alpha_j)\alpha_i \mathbf{c}_i
    \\
    \alpha_i = 1-e^{-\sigma_i\delta_i},
    \label{eq:vre-nerf}
\end{gather}

%% file: equations/mean_squared_error.tex
\begin{equation}
    L = \sum_{k} \lVert\hat{\mathbf{c}}(\mathbf{r}_k)-\hat{\mathbf{c}}_\text{gt}(\mathbf{r}_k)\rVert_2^2.
    \label{eq:mse-loss}
\end{equation}

%% file: figures/pipeline.tex
\begin{figure}[t]
    \centering
    \includegraphics[width=0.95\columnwidth]{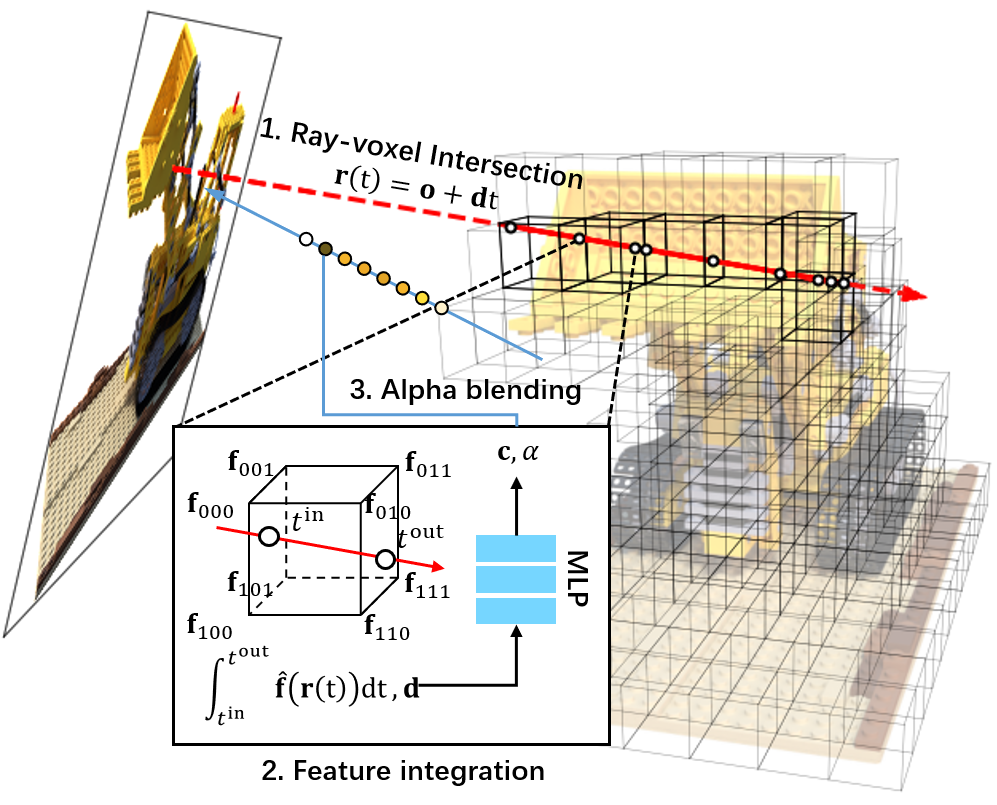}
    \caption{\textbf{Rendering pipeline overview of our DIVeR.} 
    To render a ray, we first find its intersection with voxels. For each voxel, features at its eight vertices represent a trilinear function. We integrate this trilinear function from the ray's intersection at the entry to exit, passing the result to an MLP that decodes to color and alpha values for the voxel. We obtain the final integral estimate for the ray by accumulating color and alpha values along the ray.
    }
    \label{fig:pipeline}
\end{figure}

%% file: 4-method.tex
\section{Method}
\label{sec:method}
As shown in the overall rendering pipeline (Fig.~\ref{fig:pipeline}), our DIVeR method differs from the NeRF style models in two important ways: (1) we represent the fields as a voxel grid of feature vectors $\mathbf{f}_{ijk}$, and (2) we use a decoder MLP with learnable weight $\mathbf{w}$ (Fig.~\ref{fig:architecture}) to design a deterministic integrator to estimate partial integrals of any fields of the scene. To estimate the volume rendering integral for a particular ray, we decompose it into intervals corresponding to the voxels the ray passes through. We then let each interval report an approximate estimate of the voxel's contribution and accumulate them to the rendering result. The learning of the fields is done by adjusting $\mathbf{f}_{ijk}$ and $\mathbf{w}$ to produce close approximations of the observed images.

\subsection{Voxel based deep implicit fields}
\label{subsec:deep-voxel-grid}
As in NSVF~\cite{liu2020neural}, the feature vectors are placed at vertices of the voxel grid; feature values inside each voxel are given by the trilinear interpolation of the voxel's eight corners, which yields a piecewise trilinear feature function $\mathbf{f}(\mathbf{x})$. The voxel grid can be thought of as a 3D cache of intermediate sums of NeRF's MLP, which explains why inference should be fast (Sec.~\ref{subsec:real-time-rendering}) but still can model complicated spatial behaviors compactly (Sec.~\ref{subsec:offline-rendering}). Because voxels in the empty space make no contribution to the volume rendering, the voxel grid can also be stored in a sparse representation (Sec.~\ref{subsec:inference}), which further speeds up the rendering and reduces the storage cost (Sec.~\ref{subsec:real-time-rendering}).

\input{figures/overfit}
\input{figures/implicit}

\vspace{-4mm}
\paragraph{Initializing voxel features using implicit MLP:}
If each $\mathbf{f}_{ijk}$ is trained independently and randomly initialized, our representation scheme tends to overfit during the training
(Fig.~\ref{fig:overfit}). 
This suggests that the optimization of each $\mathbf{f}_{ijk}$ should be {\em correlated}, but it is not obvious which correlation strategy should be applied.
Instead, we take an MLP that accepts the positional encoded vertex position on the voxel grid to output the feature vector at that position (the implicit MLP, with parameters $\mathbf{w}_r$; see Fig.~\ref{fig:implicit}) to correlate each feature vector implicitly. 
Although an MLP can in principle approximate any function, there is overwhelming experimental evidence that the approximated function tends to be smooth (\eg~\cite{Wizadwongsa2021NeX}), which makes it unsuitable for rendering high frequency details.
Therefore, we first train the implicit MLP to generate a reasonable initialization of $\mathbf{f}_{ijk}$ placed in the corresponding voxel grid vertex, then discard the regularization MLP and directly optimize on $\mathbf{f}_{ijk}$ explicitly. Experiments show this ‘implicit-explicit’ strategy prevents overfitting while preserving high-frequency contents.

\subsection{Feature integration}
\label{subsec:dvrf}
Intersecting a ray with the voxel grid yields a set of intervals,
which are processed separately by our integrator.
Write $(t_1^\text{in},t_1^\text{out}),\ldots,( t_n^\text{in},t_n^\text{out})$ for parameter values defining
these intervals, from eye to far end.
For interval $i$, we obtain density $\sigma_i$ and radiance $\mathbf{c}_i$ by passing
the normalized integral of $\mathbf{f}(\mathbf{x})$ along the interval to the MLP. Let $\mathbf{f}^i_1,\ldots,\mathbf{f}^i_8$ be the feature vectors at corners of the voxel the interval passes through and $\chi_1(\mathbf{x}),\ldots,\chi_8(\mathbf{x})$ be the corresponding trilinear interpolation weights, so: 
\input{equations/vre_feature_int}
Here $\mathbf{w}$ are the learnable weights of the MLP, and we incorporate viewing direction $\mathbf{d}$ to model the
view dependent effect. These approximations are accumulated into a single value of the integral by
\input{equations/vre_factorization}
(which is an approximation of Eq.~\ref{eq:vre-ori}, see the supplementary).  Notice that, if the MLP had no hidden layers, and
the integrand was a known function, we would be adjusting components of a basis function expansion of the integrand to produce the approximation. 

\input{figures/integration_comparison}
Our integrator has two advantages over MC.  First,
we get a slightly better estimate per interval (the MC estimate assumes fields inside an interval are constant; ours fits them using an MLP; see Fig.~\ref{fig:int-compare}), and this manifests in better rendering quality (Sec.~\ref{subsec:ablation}). Second, because the integrator is deterministic, the error in integral
estimates is deterministic, and so is the gradient, which may help learning; our experience has been that our
method has vanishing gradients less often than standard NeRF, and is less sensitive to the choice of learning rate.

\input{figures/architecture}
\subsection{Architecture}
\label{subsec:architecture}
We choose the feature dimension to be 32, and the voxel grid size varies according to the target image resolution.
The grid is relatively coarse and can be represented very efficiently with a sparse
representation  (Sec.~\ref{subsec:implementation}).
As shown in Fig.~\ref{fig:architecture}, we investigate two different MLP decoders: \ours32 and \ours64. Similar to \cite{mildenhall2020nerf}, we apply positional
encoding to the viewing direction $\mathbf{d}$, but we directly pass the integrated feature $\mathbf{f}$ into the MLP
without positional encoding. We use 10 bands for positional encoding in the implicit regularization MLP and 4 bands in the decoder
MLP. Because the architectures are tiny, one call of MLP takes less than 1ms, which allows MLP evaluation to happen in real time.

\subsection{Training}

We optimize $\mathbf{f}_{ijk}$, $\mathbf{w}$, and $\mathbf{w}_r$ for each scene. During a training step, we randomly sample a batch of
rays from the training set and follow the procedure described in Sec.~\ref{subsec:dvrf} to render the color, and then apply gradient descent on $\mathbf{f}_{ijk}$ and $\mathbf{w}$ using 
Eq.~\ref{eq:mse-loss}. We want the voxel grid to be sparse, and so
discourage the model from predicting background color in empty space using the regularization loss of \cite{hedman2021baking}:
\input{equations/sparity_loss}
where $\sigma_i$ denotes the $i$th accumulated density; $\lambda_s$ is the regularization weight.
In contrast to NeRF, we do not need hierarchical volume sampling because we use deterministic integration.

\vspace{-4mm}
\paragraph{Coarse to fine:} We speed up training with a coarse to fine procedure. Early in training, it is sufficient to use coarse resolution images to determine whether particular regions are empty using the culling strategy discussed in
Sec.~\ref{subsec:inference}. Based on the coarse occupancy map, we then train high resolution images and efficiently skip the empty space. When we do so, we discard the features and MLP weights trained on the coarse images (which are trained to ignore fine details).

\subsection{Inference time optimization}
\label{subsec:inference}

To avoid querying voxels that have no effect on the image (empty voxels; occluded voxels), 
we follow \cite{yu2021plenoctrees} by recording
maximum blended alpha $\prod_{j=1}^{i-1}(1-\alpha_j)\alpha_i$ for each voxel from training views, and then culling
all voxels with maximum blended alphas below the threshold $\tau_{\text{vis}}=0.01$.  This culls 98\% of voxels
on average but preserves transparent surfaces. We cull after the coarse training step (to accelerate fine-scale
training) and then again after fine-scale training.

To avoid working on voxels occluded to a certain camera view, we evaluate intervals from the eye and stop working on a ray when a transmittance
estimate $\prod_i(1-\alpha_i)$ falls below a threshold ($\tau_{\text{t}}=0.01$).
Furthermore, if an interval's alpha is below $\tau_{\text{t}}$, there is no need to evaluate color.

In contrast to other voxel based real-time applications, we do not need to convert the trained model (so there is no precision loss, \etc from discretizing the model).  While in principle, our inference time optimizations must result in
loss of accuracy, the results of Sec.~\ref{subsec:real-time-rendering} suggest that this loss is negligible.

%% file: figures/overfit.tex
\begin{figure}[t]
    \centering
    \setlength\tabcolsep{1.5pt}
    \resizebox{0.60\columnwidth}{!}{
    \begin{tabular}{cc}
         \includegraphics[width=0.4\linewidth]{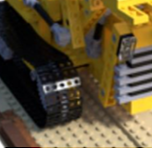}&
         \includegraphics[width=0.4\linewidth]{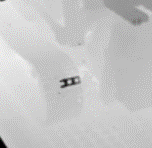}\\
    \end{tabular}
    }
    \caption{\textbf{The simple voxel model overfits:}
      Training a model with independent $\mathbf{f}_{ijk}$ has a strong tendency to overfit.  In this example,
      the model has interpreted a gloss feature as empty space. Our regularization procedure is explained in the
      text.
}
    \label{fig:overfit}
\end{figure} 

%% file: figures/implicit.tex
\begin{figure}[t]
    \centering
    \includegraphics[width=0.9\linewidth]{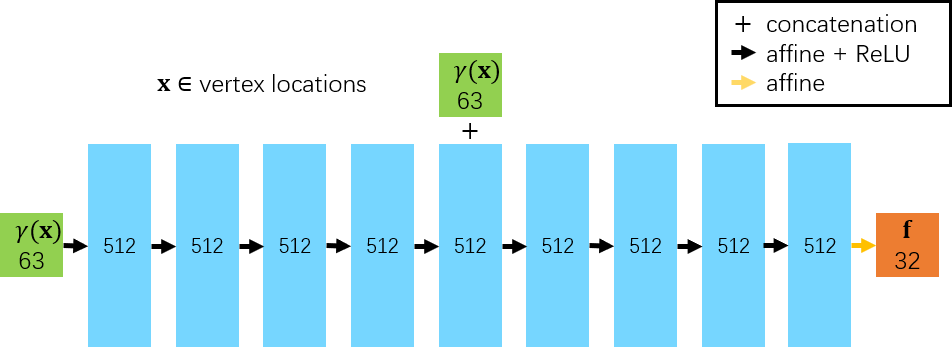}
    \caption{\textbf{The implicit MLP} generates correlated vertex features by taking a positional encoding of vertex location and producing
      a feature vector.}
    \label{fig:implicit}
\end{figure}

%% file: equations/vre_feature_int.tex
\begin{gather}
    \label{eq:feature-dec}
    (\sigma_i,\mathbf{c}_i)=\text{MLP}_\mathbf{w} (\int_{t^\text{in}_i}^{t^\text{out}_i} \hat{\mathbf{f}}(\mathbf{r}(t))dt,\mathbf{d})\\
    \begin{gathered}
    \text{where} \int_{t^\text{in}_i}^{t^\text{out}_i} \hat{\mathbf{f}}(\mathbf{r}(t))dt=\\
    \int_{t^\text{in}_i}^{t^\text{out}_i} \sum_{k=1}^{8} \mathbf{f}^i_k\frac{\chi_k(\mathbf{r}(t))}{|t^\text{out}_i-t^\text{in}_i|} dt
     = \sum_{k=1}^8 \mathbf{f}^i_k \int_{t^\text{in}_i}^{t^\text{out}_i} \frac{\chi_k(\mathbf{r}(t))}{|t^\text{out}_i-t^\text{in}_i|} dt.
    \label{eq:feature-int}
    \end{gathered}
\end{gather}

%% file: equations/vre_factorization.tex
\begin{gather}
    \label{eq:vre-factorization}
    \hat{\mathbf{c}}(\mathbf{r}) = \sum_{i=1}^{n} \prod_{j=1}^{i-1} (1-\alpha_j)\alpha_i\mathbf{c}_i\\
    \alpha_i = 1-e^{-\sigma_i}\\
    \label{eq:vre-int}
    \sigma_i = \int_{t^\text{in}_i}^{t^\text{out}_i} \sigma(\mathbf{r}(t)) dt,
    \mathbf{c}_i = \int_{t_i^\text{in}}^{t_i^\text{out}}\mathbf{c}(\mathbf{r}(t),\mathbf{d})dt
\end{gather}

%% file: figures/integration_comparison.tex
\begin{figure}[t]
    \centering
    \setlength\tabcolsep{1.0pt}
    \resizebox{0.95\columnwidth}{!}{
    \begin{tabular}{cc}
         \includegraphics[width=0.6\columnwidth]{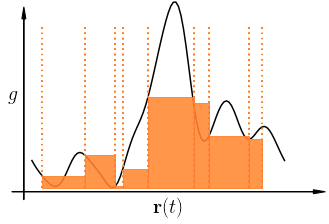}&
         \includegraphics[width=0.6\columnwidth]{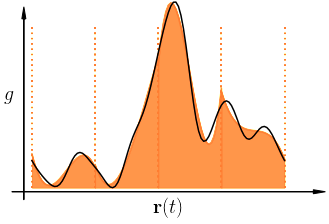}\\
         Monte Carlo & Feature integration\\ 
    \end{tabular}
    }
    \caption{\textbf{Integration strategy comparison}. Monte Carlo method fills an interval by a constant; Feature integration fits the interval with trilinear functions that are analytically integratable and blends them using an MLP.
    }
    \label{fig:int-compare}
\end{figure}

%% file: figures/architecture.tex
\begin{figure}[t]
   \centering
   \resizebox{0.75\columnwidth}{!}{
   \begin{tabular}{c}
        \includegraphics[width=0.7\linewidth]{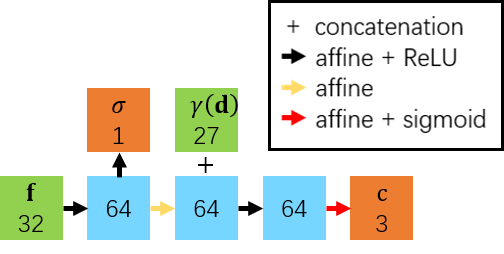}\\
        \ours64\\
        \\
        \includegraphics[width=0.7\linewidth]{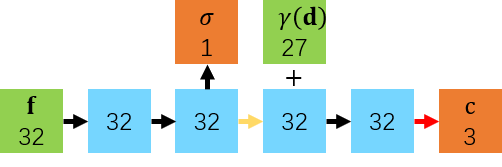}\\
        \ours32
   \end{tabular}
   }
    \caption{\textbf{Decoder architecture}. \ours64 has around
8K parameters; \ours32 has around 4k parameters. Both MLPs take integrated features and positional encoded viewing direction as inputs and output corresponding integrated density and color.}
\label{fig:architecture}
\end{figure}

%% file: equations/sparity_loss.tex
\begin{equation}
    L_{\text{sparsity}} = \lambda_{s}\sum_i \log(1+\frac{\sigma_i^2}{0.5}),
\end{equation}

%% file: 5-experiments.tex
\section{Experiments}
\label{sec:experiments}
We evaluate using both the offline rendering task (FPS$\leq$20) and the real-time rendering task (FPS$>$20). We
use the NeRF-synthetic dataset \cite{mildenhall2020nerf} (synthetic images of size $800 \times 800$ with
camera poses);  a subset of the Tanks and Temples dataset \cite{10.1145/3072959.3073599} and the BlendedMVS dataset
\cite{yao2020blendedmvs} (chosen by NSVF authors \cite{liu2020neural}). Tanks and Temples images are $1920 \times 1080$;
BlendedMVS images are $768 \times 576$. Backgrounds in both datasets are cropped by NSVF.
The qualitative results of our experiments can be seen in Fig.~\ref{fig:offline-qual2} and Fig.~\ref{fig:offline-qual}.  In all quantitative
measurements, we mark the best result by bold font and second best by italic font with underline. 

\subsection{Implementation detail}
\label{subsec:implementation}
\paragraph{Training:}
We use PyTorch~\cite{NEURIPS2019_9015} for network optimization and customized CUDA kernels to accelerate ray-voxel
intersection. 
For high resolution image training, both implicit and explicit models use a voxel grid of size $256^3$ for NeRF-synthetic and
BlendedMVS, and $320^3$ for Tanks and Temples. 
For coarse model training, we take the voxel grid and images at 1/4 of the fine model scale.
We follow NSVF's~\cite{liu2020neural} strategy to sample
rays from the training set, and we choose a batch size of 1024 pixels for coarse training, 6144 for fine training of Tanks and Temples, and 8192 for fine
training of other datasets. The coarse model is trained for 5 epochs first explicitly, and then we train the model with implicit MLP until the validation loss has
almost converged. Finally, we train the explicit grid initialized from the implicit model and stop the training when
the total training time reaches 3 days. In total, the peak GPU memory usage is around 40GB. We use the
Adam~\cite{Kingma2015AdamAM} optimizer with a learning rate of 5e-4 for the fine model, 1e-3 for the coarse model, and $\lambda_s=$1e-5 in the sparsity regularization loss. 

\input{tables/offline_quality}
\vspace{-4mm}
\paragraph{Real-time application:}
Our real-time application is implemented by using CUDA and Python, with all the operations being parallelized per image
pixel. For each frame and each pixel, ray marching finds a fixed number of hits on the voxel
grid; the MLP is evaluated for each hit, and the result is then blended to the image buffer. This sequence is repeated
until the ray termination criteria is reached (Sec.~\ref{subsec:inference}). 

\vspace{-4mm}
\paragraph{Storage:}
Because the voxel grid is sparse, we need to store only: indices and values of feature vectors for non-empty voxels;
a binary occupancy mask; and the MLP weight. At inference, we keep feature vectors in a 1D array
and then build a dense 3D array that stores the indices to the specific feature value, thereby reducing GPU memory
demand without much sacrifice of performance. 

\subsection{Offline rendering}
\label{subsec:offline-rendering}
We evaluate the offline model by measuring the similarity between rendered and ground
truth images using PSNR, SSIM \cite{1284395}, and LPIPS \cite{zhang2018perceptual}. 
We use our \ours64 model for all scenes.

\input{figures/offline-quality}

\vspace{-4mm}
\paragraph{Baselines:} We compare with original NeRF \cite{mildenhall2020nerf}; the reimplementation in Jax
\cite{jaxnerf2020github};  AutoInt~\cite{autoint2021}; and NSVF~\cite{liu2020neural} (which uses similar voxel grid features).
Pre-trained models from real-time NeRF variants produce good rendering quality but
are trained and evaluated on very large computational resources (for example, JaxNeRF+~\cite{hedman2021baking}
doubles the feature size of NeRF's MLP and takes 5 times more samples for volume
rendering, which is impractical for evaluation on a standard GPU).  Therefore, we exclude them from the baseline models.

\vspace{-4mm}
\paragraph{Results:}
\text{\ours} rendering quality is comparable with other offline baselines, while its architecture is much simpler
(Tab.~\ref{tab:offline}). Our PSNR is only slightly worse than that of NSVF on Tanks and Temples;
but we use a much simpler decoder MLP.
\input{figures/offline-quality-2}
\input{tables/realtime-quality}
\input{tables/realtime-efficiency}

\subsection{Real-time rendering}
\label{subsec:real-time-rendering}
For the real-time rendering task, we use our \ours32 model with the inference time optimization described in
Sec.~\ref{subsec:inference}. Besides rendering quality, we show the inference time efficiency by running all the models
on a GTX1080 GPU and recording their FPS and GPU memory usage. To compare the compactness of the architecture, we report
the average memory usage for storing a scene. As most real-time models are converted from some pre-trained models, we
also show the rendering quality of those models and compare the precision loss after the conversion. For the models that
have variants based on quality-speed trade-off, we report their variants with the best rendering quality. 

\vspace{-4mm}
\paragraph{Baselines:}
For our real-time rendering baselines, we compare with PlenOctrees~\cite{yu2021plenoctrees},
SNeRG~\cite{hedman2021baking}, FastNeRF~\cite{garbin2021fastnerf}, and KiloNeRF~\cite{reiser2021kilonerf}. Since SNeRG
did not provide their pre-trained models, we directly report the measurements from their paper. For the same reason, we
report only the rendering quality for the FastNeRF as reported in their paper. For KiloNeRF, we measure the performance
of those scenes for which their model was made available (chair, lego, and ship). 

\vspace{-4mm}
\paragraph{Results:}
Our rendering quality is either best (Tab.~\ref{tab:realtime-perf}) or second best (Tab.~\ref{tab:realtime-qual}),
but our method achieves very high frame rates for very small models.
All other methods 
must (1) convert to achieve a real-time form and then (2) fine-tune to recover precision loss after conversion. Fine-tuning is crucial for these models; for example, if SNeRG is not fine-tuned, its PSNR degrades dramatically to 26.68. In contrast, our model is evaluated as trained without conversion or fine-tuning.
Early ray termination (Sec.~\ref{subsec:inference}) causes the mild degradation in quality observed in our real-time methods.

\subsection{Ablation study}
\label{subsec:ablation}
\input{tables/network_ablation}
\input{figures/architecture_ablation}
\paragraph{Architecture:}
We perform all our ablation studies on the NeRF-synthetic dataset. In Tab. \ref{tab:network-abl}, we show the performance
trade-off between different network architectures. Without any real-time optimization, our model still runs faster than
regular NeRF that takes minutes to run a single frame; if we use a smaller decoder for speed, there is a minor
loss of quality but speed doubles (because \ours32 uses half as many registers as \ours64, allowing more threads to
run in the CUDA kernel). Further economy with acceptable PSNR can be obtained by reducing the voxel grid size
(compare \ours32 at 128 voxels yielding 30.42 PSNR for an about 12MB model to PlenOctree's variant with 30.7 PSNR, about 400MB). Fig.~\ref{fig:architecture-alb} compares different MLP sizes and voxel grid sizes qualitatively.

\input{figures/scene_manipulation}
\input{tables/training_ablation}

\vspace{-5mm}
\paragraph{Training strategy:}
Tab.~\ref{tab:training-abl} shows the effect of different training strategies on the lego scene trained with our \ours64 model.  The deterministic integrator is important: using a random integrator (implemented with sampling strategy
of NSVF~\cite{liu2020neural}) in train and test causes a notable loss of quality. The implicit-explicit training strategy is important: replacing it  either with a pure implicit model or with no implicit MLP initialization (compare Fig.~\ref{fig:overfit}) results in a less significant loss of quality.  A
lower precision representation of the feature vectors (trained with a tanh mapping; converted to unit8) results in a minor loss of quality, but the model size is reduced by a factor of 3.

\vspace{-4mm}
\paragraph{Editability:} The voxel based representation allows us to perform some basic scene manipulations. We can  composite scenes by blending their voxel grids and then using the corresponding decoder for rendering. Because feature vectors incorporate high level information of the local appearance, we can  extract the segmentation of an object from a selected area by using k-mean clustering on the feature vectors, which allows us to swap objects without noticeable artifacts. Fig.~\ref{fig:scene-manipulation} shows some examples.

%% file: tables/offline_quality.tex
\begin{table}
\centering
\resizebox{\columnwidth}{!}{
\setlength\tabcolsep{3 pt}
\begin{tabular}{clccc}
\toprule
    & Method & PSNR $\uparrow$ & SSIM $\uparrow$ & LPIPS $\downarrow$ \\\midrule
\multirow{4}{*}{NeRF-Synthetic}    
    & NeRF~\cite{mildenhall2020nerf}    & 31.00    & 0.947 & 0.081\\ 
    & JaxNeRF~\cite{jaxnerf2020github} & 31.65 & 0.952 & 0.051\\
    & AutoInt~\cite{autoint2021} & 25.55 & 0.911 & 0.170\\
    & NSVF~\cite{liu2020neural}    & \textbl{31.74} & \textbl{0.953} & \textbl{0.047}\\
    & \ours64 & \textbf{32.32} & \textbf{0.960}  & \textbf{0.032}\\
    \midrule
\multirow{4}{*}{BlendedMVS}    
    & NeRF~\cite{mildenhall2020nerf}    & 24.15    & 0.828 & 0.192\\ 
    & JaxNeRF~\cite{jaxnerf2020github} & - & - & -\\
    & AutoInt~\cite{autoint2021} & - & - & -\\
    & NSVF~\cite{liu2020neural}    & \textbl{26.90} & \textbl{0.898} & \textbl{0.113}\\
    & \ours64  & \textbf{27.25} & \textbf{0.910}  & \textbf{0.073}\\ 
    \midrule
    
\multirow{4}{*}{Tanks \& Temples}    
    & NeRF~\cite{mildenhall2020nerf}    & 25.78    & 0.864 & 0.198\\ 
    & JaxNeRF~\cite{jaxnerf2020github} & 27.94 & 0.904 & 0.168\\
    & AutoInt~\cite{autoint2021} & - & - & - \\
    & NSVF~\cite{liu2020neural}    & \textbf{28.40} & \textbl{0.900} & \textbl{0.153}\\
    & \ours64 & \textbl{28.18} & \textbf{0.912}  & \textbf{0.116}\\
    \bottomrule
\end{tabular}
}
\caption{\textbf{Quantitative results on different benchmarks} show ours (\ours64) is overall the best compared with NeRF and its variant for offline rendering. `-' means no publicly available results. (\textbf{Best}; \textbl{Second best}).
}
\label{tab:offline}
\end{table}

%% file: figures/offline-quality.tex
\begin{figure*}[t]
    \centering
    \setlength\tabcolsep{1.0pt}
    \resizebox{0.95\textwidth}{!}{
    \begin{tabular}{cccc}
        \multicolumn{4}{c}{\textbf{NeRF-Synthetic}}\\
         NeRF~\cite{mildenhall2020nerf} & PlenOctrees~\cite{yu2021plenoctrees} & \ours32(RT) & Ground Truth \\
         \includegraphics[width=0.25\linewidth]{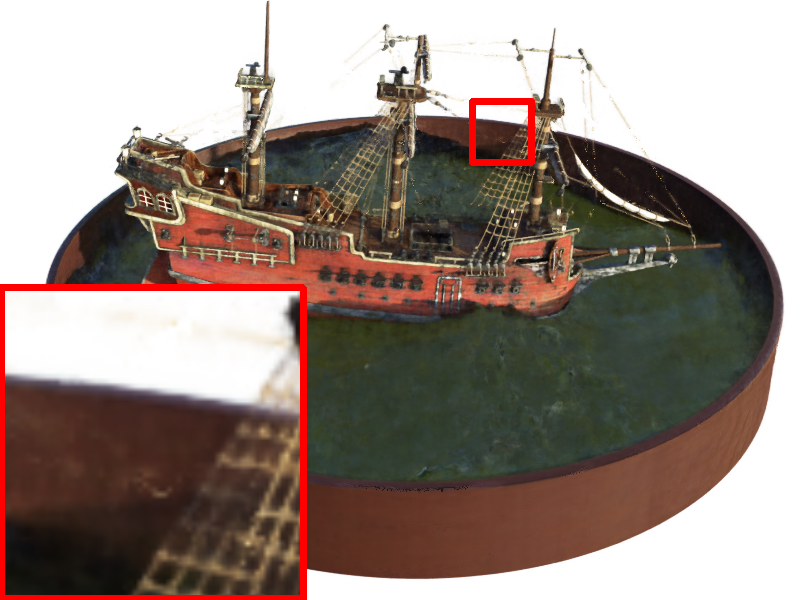}&
         \includegraphics[width=0.25\linewidth]{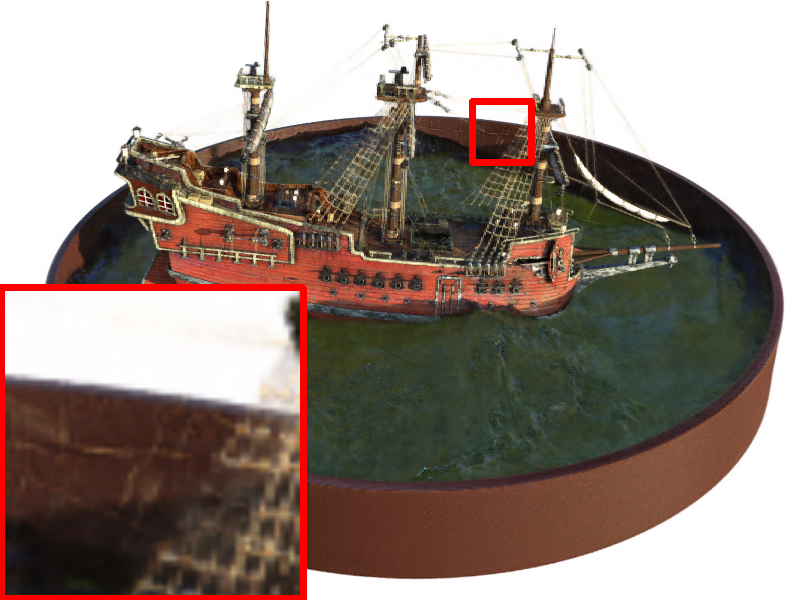}&
         \includegraphics[width=0.25\linewidth]{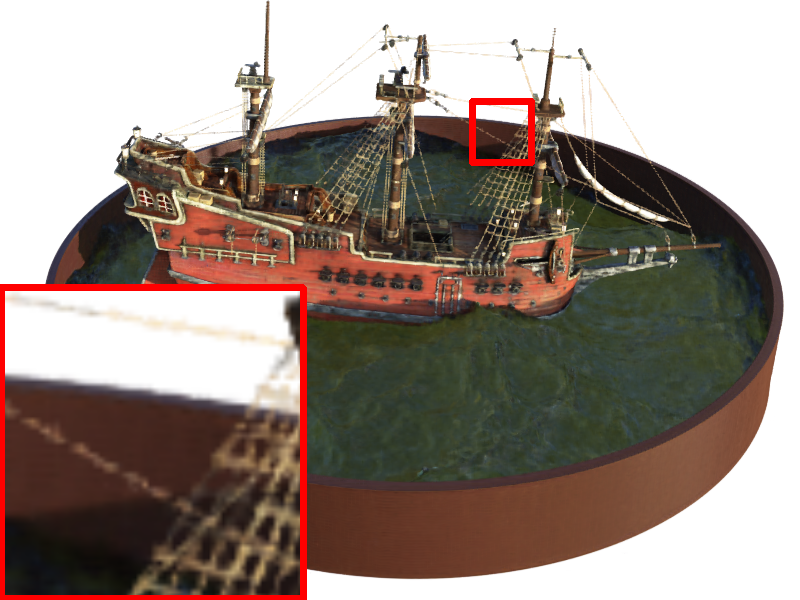}&
         \includegraphics[width=0.25\linewidth]{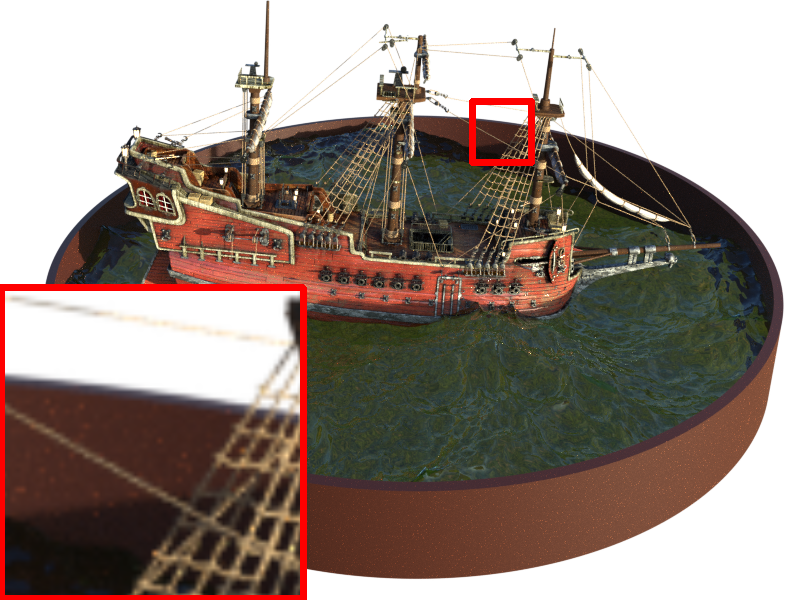}\\
         \includegraphics[width=0.25\linewidth]{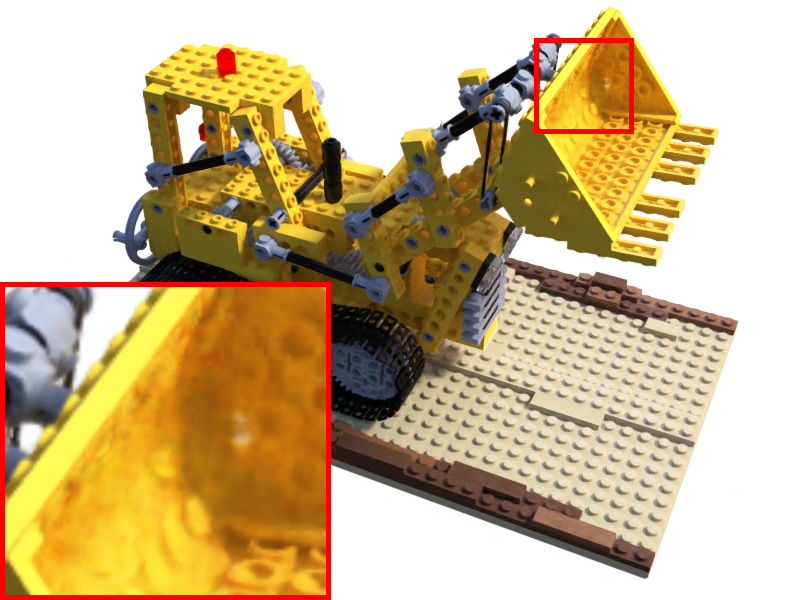}&
         \includegraphics[width=0.25\linewidth]{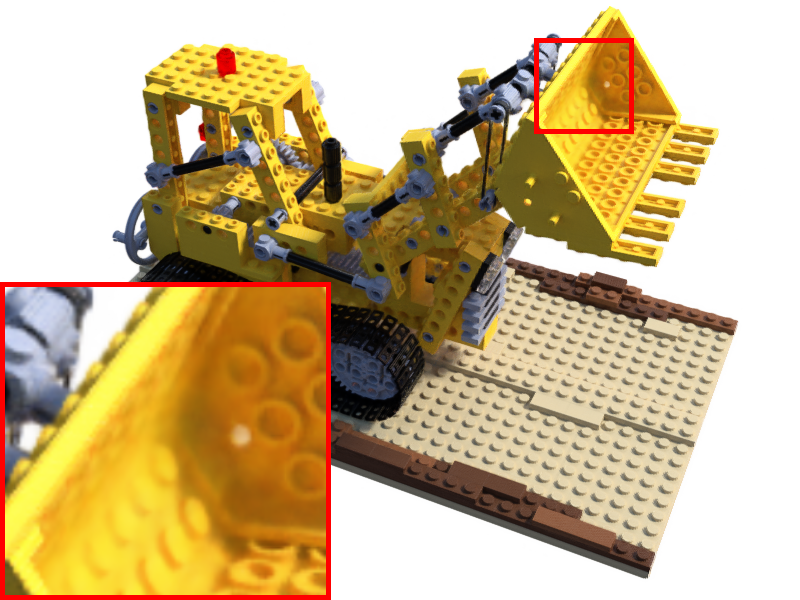}&
         \includegraphics[width=0.25\linewidth]{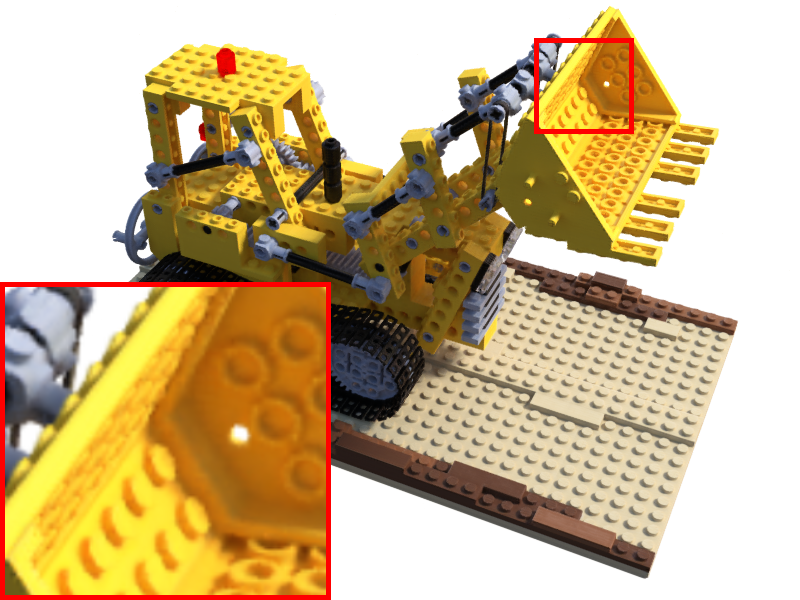}&
         \includegraphics[width=0.25\linewidth]{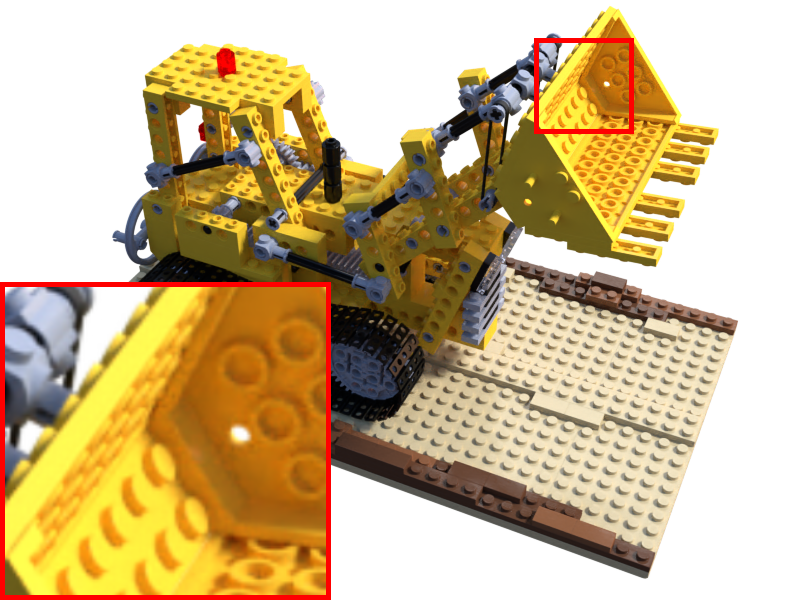}\\
         \includegraphics[width=0.25\linewidth]{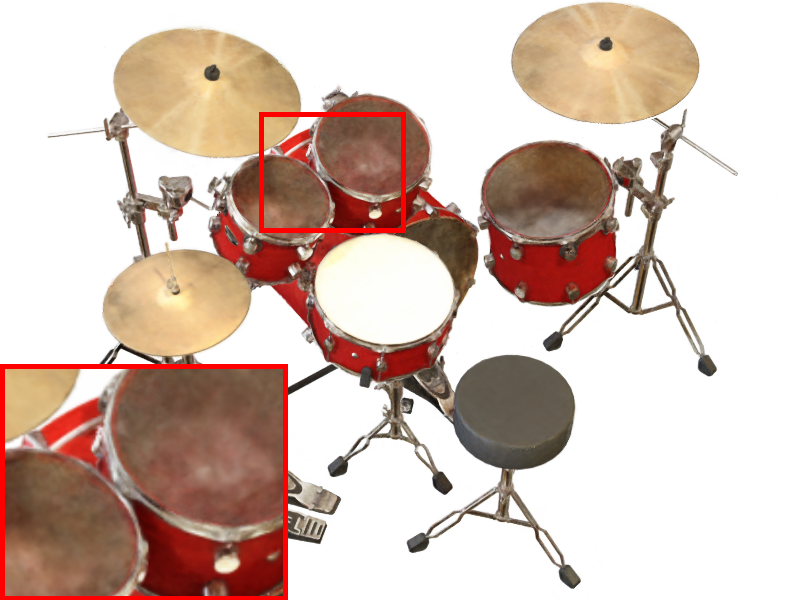}&
         \includegraphics[width=0.25\linewidth]{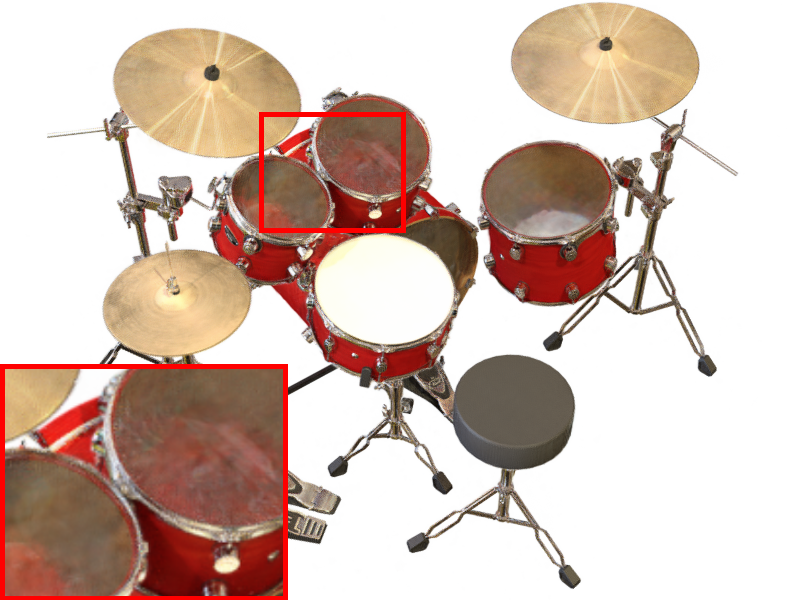}&
         \includegraphics[width=0.25\linewidth]{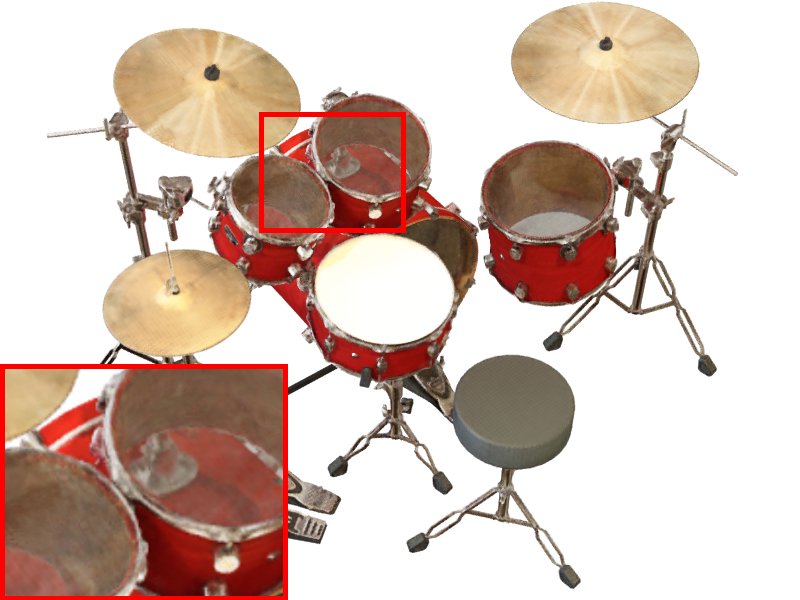}&
         \includegraphics[width=0.25\linewidth]{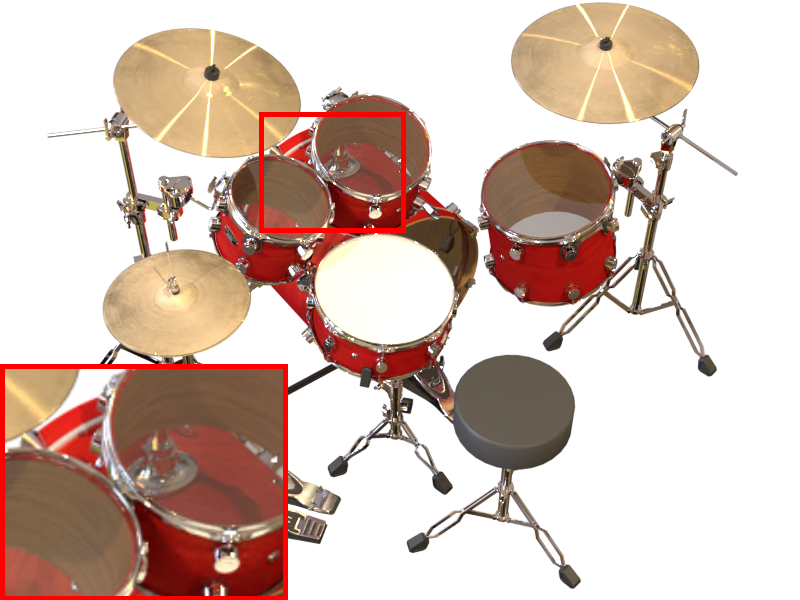}
    \end{tabular}
    }
    \caption{\textbf{Qualitative rendering results} show that
      our method successfully models fine and translucent structures (shrouds and ratlines on the ship; studs on lego;
      drumhead) which NeRF and PlenOctrees find hard. Our method is slightly slower than PlenOctrees (real-time) but much smaller; NeRF runs offline.}
    \label{fig:offline-qual2}
\end{figure*}

%% file: figures/offline-quality-2.tex
\begin{figure*}[t]
    \centering
    \setlength\tabcolsep{1.0pt}
    \resizebox{0.95\textwidth}{!}{
    \begin{tabular}{cccc}
         \multicolumn{2}{c}{\textbf{Tanks and Temples}}&
         \multicolumn{2}{c}{\textbf{BlendedMVS}}\\
         \ours64 & Ground Truth & \ours64 & Ground Truth\\
         \includegraphics[width=0.25\linewidth]{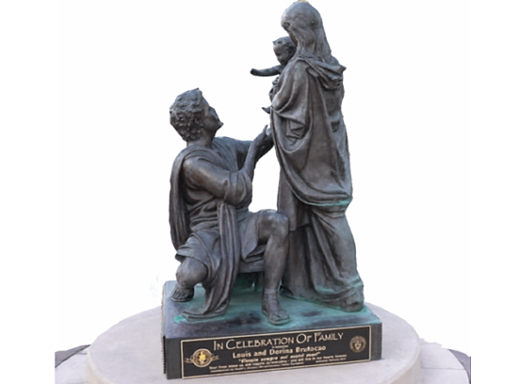}&
         \includegraphics[width=0.25\linewidth]{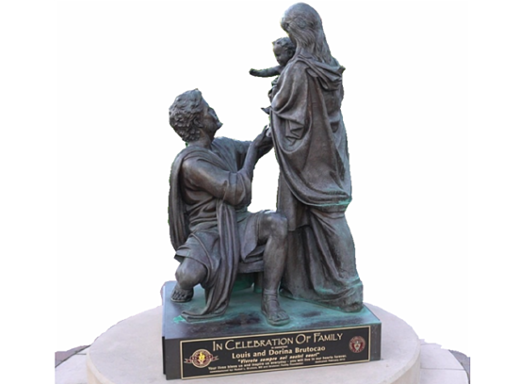}&
         \includegraphics[width=0.25\linewidth]{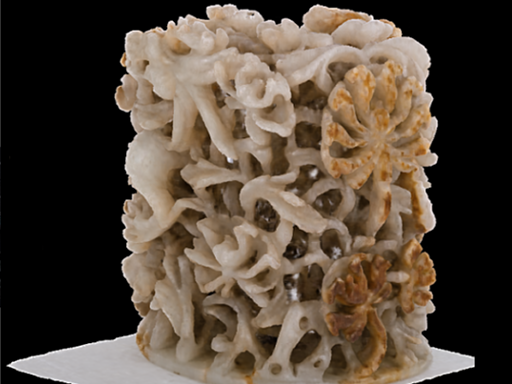}&
         \includegraphics[width=0.25\linewidth]{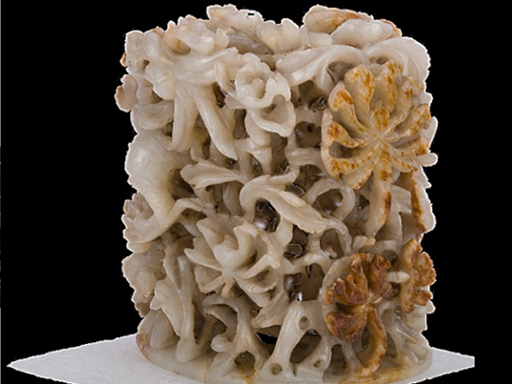}\\
         \includegraphics[width=0.25\linewidth]{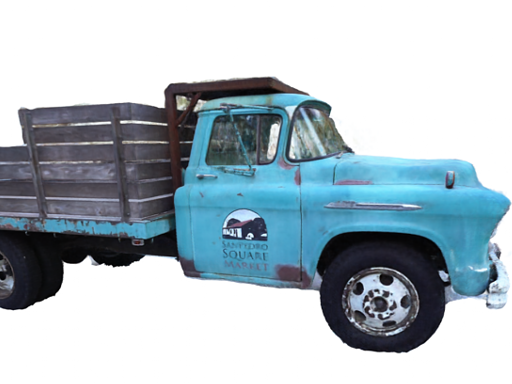}&
         \includegraphics[width=0.25\linewidth]{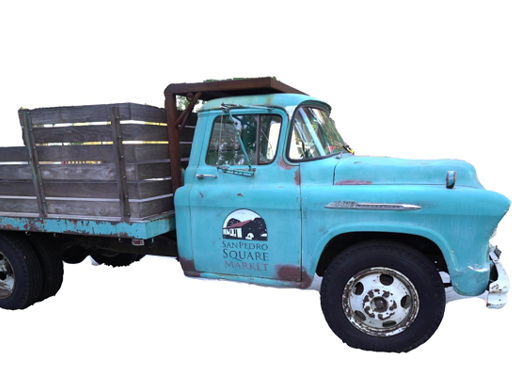}&
         \includegraphics[width=0.25\linewidth]{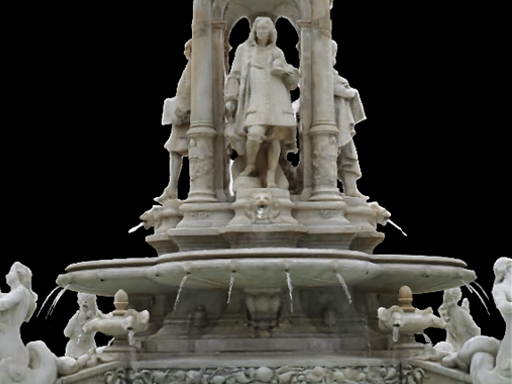}&
         \includegraphics[width=0.25\linewidth]{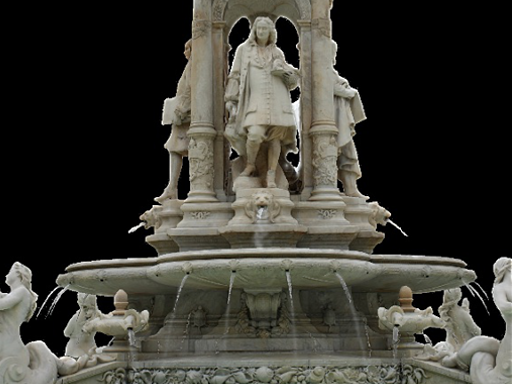}\\
    \end{tabular}
    }
    \caption{\textbf{Qualitative rendering results} of Tanks and Temples and BlendedMVS data show our method successfully models real-world fine structures. 
    }
    \label{fig:offline-qual}
\end{figure*}

%% file: tables/realtime-quality.tex
\begin{table}
\centering
\begin{tabular}{l c c c}
     \toprule
     Method & PSNR $\uparrow$ & SSIM $\uparrow$ & LPIPS $\downarrow$ \\ \midrule 
     NeRF-SH~\cite{yu2021plenoctrees}  &  31.57   &  0.952 & 0.063 \\
     JaxNeRF+~\cite{hedman2021baking} &  \textbf{33.00} & \textbf{0.962} & \textbl{0.038} \\
     NeRF~\cite{mildenhall2020nerf} &  31.00 & 0.947 & 0.081 \\
     \ours32   &  \textbl{32.16} & \textbl{0.958} & \textbf{0.032} \\
     \bottomrule
\end{tabular}

    \caption{\textbf{Image quality comparisons to real-time pre-trained models} on NeRF-synthetic strongly support our method (almost as good as
  JaxNeRF+). In Tab.~\ref{tab:realtime-perf}, we show the performance of their corresponding baked real-time applications.}
\label{tab:realtime-qual}
\end{table}

%% file: tables/realtime-efficiency.tex
\begin{table}
\centering
\resizebox{\columnwidth}{!}{
\setlength\tabcolsep{2.5 pt}
\begin{tabular}{l c c c c c c}
    \toprule
     Method & PSNR $\uparrow$ & SSIM $\uparrow$ & LPIPS $\downarrow$ & 
              FPS $\uparrow$ & MB $\downarrow$ & GPU GB $\downarrow$\\
     \midrule
     PlenOctrees~\cite{yu2021plenoctrees}     & \textbl{31.71} & \textbf{0.958} & 0.053 &
                       \textbl{76$\pm$66}   & 1930  & 1.65$\pm$1.09\\
     SNeRG~\cite{hedman2021baking}           & 30.38 & \textbl{0.950} & 0.050 &
                       \textbf{98$\pm$37}   &  \textbl{84}   & 1.73$\pm$1.48\\
     FastNeRF~\cite{garbin2021fastnerf}       &  29.97  & 0.941 & 0.053 & - & - & - \\
     KiloNeRF~\cite{reiser2021kilonerf}    & 31.00 & 0.\textbl{950} & \textbf{0.030} &
                   28$\pm$12    &  161  & \textbl{1.68$\pm$0.27}\\
     \ours32(RT)      & \textbf{32.12} & \textbf{0.958} & \textbl{0.033}&
                   47$\pm$20    &  \textbf{68}   & \textbf{1.07$\pm$0.06}\\
    \bottomrule
\end{tabular}
}
\caption{\textbf{Comparisons to other real-time variants} on NeRF-synthetic show that our method produces small models with very low GPU demand and renders very fast with very strong image quality metrics. PlenOctrees is baked from NeRF-SH; SNeRG is baked from
  JaxNeRF+; KiloNeRF and FastNeRF directly convert from the original
  NeRF. Performance measurements of FastNeRF are not publicly available.
} 
\label{tab:realtime-perf}
\end{table}

%% file: tables/network_ablation.tex
\begin{table}
\centering

\begin{tabular}{c l l c c c }
\toprule
$N$ & Decoder & RT & PSNR $\uparrow$ & FPS $\uparrow$ & MB $\downarrow$ \\
\midrule
256 & \ours64 & No  & \textbf{32.32} & 0.62 & \textbl{62} \\
256 & \ours32 & No & 32.16          & 0.62 & 68 \\
128 & \ours64 & No & 30.72          & 0.62 & \textbf{12} \\
128 & \ours32 & No & 30.53          & 0.62 & \textbf{12} \\

256 & \ours64 & Yes & \textbl{32.30}    & 26$\pm$9  & \textbl{62} \\
256 & \ours32 & Yes & 32.12          & \textbl{47$\pm$20}  & 68 \\
128 & \ours64 & Yes & 30.64          & 37$\pm$20  & \textbf{12} \\
128 & \ours32 & Yes & 30.52          & \textbf{82$\pm$37}  & \textbf{12} \\
\bottomrule
\end{tabular}
\caption{\textbf{Ablation study on network architecture} shows that real-time optimization (RT) and smaller MLPs (Decoder) involve minimal loss of rendering quality but huge speedups; going to a coarser grid ($N$) involves a larger loss of quality, for further very large speedup and improvement in model size.
}
\label{tab:network-abl}
\end{table}

%% file: figures/architecture_ablation.tex
\begin{figure}[t]
   \centering
   \setlength\tabcolsep{1.5pt}
   \resizebox{0.98\columnwidth}{!}{
   \begin{tabular}{cccc}
        \includegraphics[width=0.3\columnwidth]{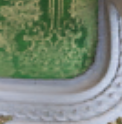} &
        \includegraphics[width=0.3\columnwidth]{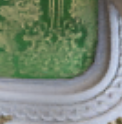} &
        \includegraphics[width=0.3\columnwidth]{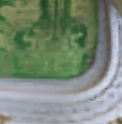} &
        \includegraphics[width=0.3\columnwidth]{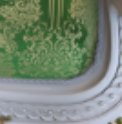}\\
        \ours64 & \ours32 & \ours64 & \multirow{2}{*}{Ground Truth} \\
        $N=256$ & $N=256$ & $N=128$ 
   \end{tabular}
   }
    \caption{\textbf{Qualitative comparison of different architectures}
        shows larger voxel grid can better model the fine texture on the chair surface; switching to a smaller MLP does not affect the quality too much. 
    }
\label{fig:architecture-alb}
\end{figure}

%% file: figures/scene_manipulation.tex
\begin{figure*}[ht!]
   \centering
   \setlength\tabcolsep{1.5pt}
   \resizebox{0.95\textwidth}{!}{
   \begin{tabular}{ccccccc}
       \multicolumn{3}{c}{
        \includegraphics[width=0.4\textwidth]{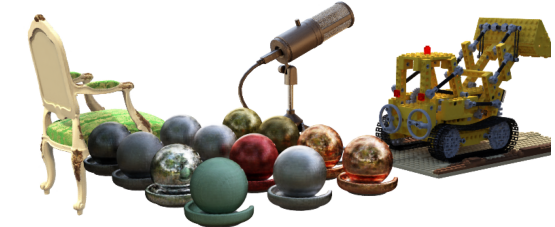}}&&
        \includegraphics[width=0.2\textwidth]{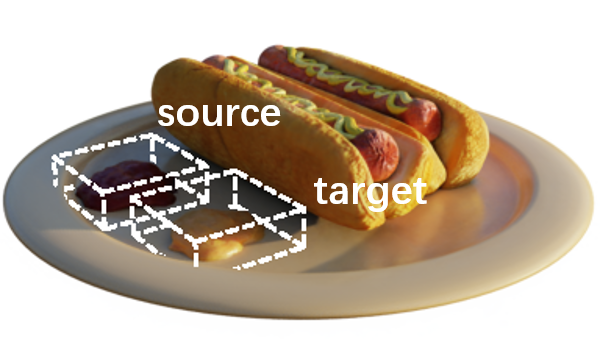}&
        \includegraphics[width=0.2\textwidth]{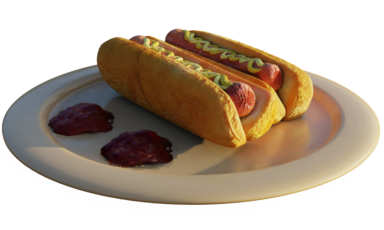}&
        \includegraphics[width=0.2\textwidth]{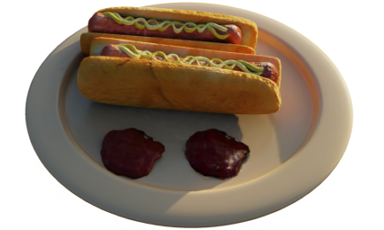}\\
        \multicolumn{3}{c}{Scene composition}&&
        \multicolumn{3}{c}{Object swapping}\\
   \end{tabular}
   }
    \caption{Our representation admits 
    useful \textbf{editing} in a straightforward way.  {\bf Left:} shows a composite of pretrained models, obtained by blending voxel grids.  {\bf Right:} the mustard on the plate (first hot-dog) is replaced with ketchup (second hot-dog); this edit is preserved under view change (third hot-dog).
    }
\label{fig:scene-manipulation}
\end{figure*}

%% file: tables/training_ablation.tex
\begin{table}
\centering

\setlength\tabcolsep{3 pt}
\begin{tabular}{l l l c c}
\toprule
Integrator & Regularization   & Data type & 
PSNR $\uparrow$  & MB $\downarrow$\\
\midrule
Det & Im-Ex & float32 &\textbf{35.52} & \textbl{64}\\
Rand  & Im-Ex & float32 & 33.89       & 67 \\
Det & Im-Ex & uint8 & \textbl{35.44} & \textbf{19}\\
Det & Im &  float32 & 34.69  & \textbl{64}\\
Det & Ex & float32 & 34.02  & \textbl{64}\\
\bottomrule
\end{tabular}
\caption{\textbf{Ablation study on training strategy}. Implicit (Im) MLP initialization with explicit (Ex) training gives the best rendering quality. Mapping feature vectors with tanh allows features to be stored more efficiently (uint8) with acceptable loss of rendering qualities.}
\label{tab:training-abl}
\end{table}

%% file: 6-conclusion.tex
\section{Limitations}
\label{sec:conclusion}

\input{figures/fail_drifting}
\input{figures/fail_lighting}

Training NeRF-like representations is expensive, and our method does not speed up training in any
natural way.  Aliasing error in a deterministic integrator tends to be patterned,
whereas a stochastic integrator breaks it up~\cite{cookstoch}.  In turn, rays near tangent to voxels
or accumulated error in the intersection routine can cause problems (Fig.~\ref{fig:fail-drifting}). 
A mixed stochastic-deterministic method (say, jittering voxel positions) may help.
Our method, like NeRF,  can fail to model view dependent effects correctly (Fig.~\ref{fig:fail-lighting}); more
physical modeling might help. Our method does not currently apply to unbounded scenes, and our editing abilities are currently quite limited. 

%% file: figures/fail_drifting.tex
\begin{figure}[t]
   \centering
   \setlength\tabcolsep{1.5pt}
   \resizebox{0.98\columnwidth}{!}{
   \begin{tabular}{cccc}
        \includegraphics[width=0.3\columnwidth]{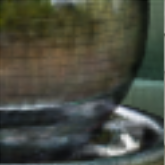} &
        \includegraphics[width=0.3\columnwidth]{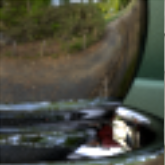} &
        \includegraphics[width=0.3\columnwidth]{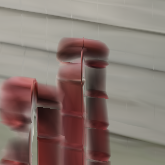} &
        \includegraphics[width=0.3\columnwidth]{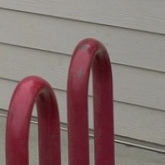}\\
        Ours & Ground Truth & Ours & Ground Truth
   \end{tabular}
   }
    \caption{\textbf{Intersection errors} can cause aliasing problems, typically for 
    structures near the scale of the voxel grid and rays near tangent to voxel faces.  
    Worse, our current intersection routine, while efficient, can
    accumulate intersection errors, occasionally producing blocky aliasing artifacts at some view directions and scales (the scale of the blocky artifacts in each case is close to the scale of the voxel grid).}
\label{fig:fail-drifting}
\end{figure}

%% file: figures/fail_lighting.tex
\begin{figure}[t]
   \centering
   \setlength\tabcolsep{1.5pt}
   \resizebox{0.735\columnwidth}{!}{
   \begin{tabular}{ccc}
        \includegraphics[width=0.3\columnwidth]{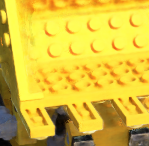} &
        \includegraphics[width=0.3\columnwidth]{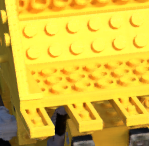} &
        \includegraphics[width=0.3\columnwidth]{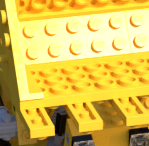}\\
        NeRF~\cite{mildenhall2020nerf} &
        Ours & Ground Truth
   \end{tabular}
   }
    \caption{\textbf{NeRF variants fail to extrapolate view dependent effect} that is unseen from the training set. Neither ours nor other NeRF variants are able to model the reflection on the shovel correctly.}
\label{fig:fail-lighting}
\end{figure}

%% file: appendix/implementation_details.tex
\section{Additional Implementation Details}
\label{sec:additional-implementation}

\subsection{Volume rendering approximation}
\label{subsec:volume-rendering-approximation}
Given a ray $\mathbf{r}(t)=\mathbf{x}+\mathbf{d}t$ and its intersection with the voxel grid $(t_1^\text{in},t_1^\text{out}),\ldots,( t_n^\text{in},t_n^\text{out})$ for parameter values from eye to far end and using the notation from the main text, the volume rendering equation can be decomposed to:
\begin{equation}
\begin{split}
    \hat{\mathbf{c}}(\mathbf{r})&=\int_0^{\infty}e^{-\int_0^t \sigma(\mathbf{r}(\tau))d\tau}\sigma(\mathbf{r}(t))\mathbf{c}(\mathbf{r}(t),\mathbf{d})dt.\\
    &=\sum_{i=1}^n \int_{t_i^\text{in}}^{t_i^\text{out}} e^{-\int_0^{t}\sigma(\mathbf{r}(\tau))d\tau} \sigma(\mathbf{r}(t)) \mathbf{c}(\mathbf{r}(t)) dt\\
    &= \sum_{i=1}^n e^{-\sum_{j=1}^{i-1}\int_{t_j^\text{in}}^{t_j^\text{out}}\sigma(\mathbf{r}(t))dt}\\
    &\times 
    \int_{t_i^\text{in}}^{t_i^\text{out}} e^{-\int_{t_i^\text{in}}^{t}\sigma(\mathbf{r}(\tau))d\tau} \sigma(\mathbf{r}(t)) \mathbf{c}(\mathbf{r}(t)) dt\\
    &=  \sum_{i=1}^n \prod_{j=1}^{i-1}(1-\alpha_j)\\
    &\times \int_{t_i^\text{in}}^{t_i^\text{out}} e^{-\int_{t_i^\text{in}}^{t}\sigma(\mathbf{r}(\tau))d\tau}
        \sigma(\mathbf{r}(t)) \mathbf{c}(\mathbf{r}(t)) dt.
\end{split}
\end{equation}
By applying Holder's inequality to the nested integration, we have:
\begin{equation}
    \begin{split}
    \int_{t_i^\text{in}}^{t_i^\text{out}} &e^{-\int_{t_i^\text{in}}^{t}\sigma(\mathbf{r}(\tau))d\tau}
        \sigma(\mathbf{r}(t)) \mathbf{c}(\mathbf{r}(t)) dt\\
        &\leq \int_{t_i^\text{in}}^{t_i^\text{out}} e^{-\int_{t_i^\text{in}}^t\sigma(\mathbf{r}(\tau))d\tau}
        \sigma(\mathbf{r}(t))dt \int_{t_i^\text{in}}^{t_i^\text{out}}\mathbf{c}(\mathbf{r}(t)) dt\\
        &= \mathbf{c}_i\int_{t_i^\text{in}}^{t_i^\text{out}} e^{-\int_{t_i^\text{in}}^t\sigma(\mathbf{r}(\tau))d\tau}
        d(\int_{t_i^\text{in}}^t\sigma(\mathbf{r}(\tau))d\tau)\\
        &= \mathbf{c}_i (-e^{-\int_{t_i^\text{in}}^t\sigma(\mathbf{r}(\tau))d\tau}|_{t=t_i^\text{in}}^{t_i^\text{out}})\\
        &= \mathbf{c}_i (1-e^{-\int_{t_i^\text{in}}^{t_i^\text{out}}\sigma(\mathbf{r}(\tau))d\tau})\\
        &= \alpha_i \mathbf{c}_i.
    \end{split}
\end{equation}
Therefore, we have:
\begin{equation}
    \hat{\mathbf{c}}(\mathbf{r})\leq \sum_{i=1}^n \prod_{j=1}^{i-1}(1-\alpha_j)\alpha_i \mathbf{c}_i.
\end{equation}

\subsection{Feature integration}
\label{subsec:feature-integration}
Assume the size of every voxel is $1\times1\times1$ and a voxel has feature vectors $\mathbf{f}_1,\ldots,\mathbf{f}_8$ placed at its eight corners. In the voxel's local coordinate system, the feature value inside the voxel at $\mathbf{x}=(x,y,z)$ is then given by
\begin{equation}
    \mathbf{f}(x,y,z) = \sum_{k=1}^8 \mathbf{f}_k \chi_k(x,y,z),\\
\end{equation}
where:
\begin{equation}
\begin{cases}
    \chi_8(x,y,z)&=xyz\\
    \chi_7(x,y,z)&=(1-x)yz\\
    \chi_6(x,y,z)&=x(1-y)z\\
    \chi_5(x,y,z)&=(1-x)(1-y)z\\
    \chi_4(x,y,z)&=xy(1-z)\\
    \chi_3(x,y,z)&=(1-x)y(1-z)\\    
    \chi_2(x,y,z)&=x(1-y)(1-z)\\
    \chi_1(x,y,z)&=(1-x)(1-y)(1-z)\\
\end{cases}.
\label{eq:trilinear}
\end{equation}
Let $\mathbf{x}_0=(x_0,y_0,z_0), \mathbf{x}_1=(x_1,y_1,z_1)$ be the voxel's intersection with a ray at entry and exit. It defines a ray segment $\mathbf{x}(t)$:
\begin{equation}
\begin{split}
    \mathbf{x}(t)&=(x(t),y(t),z(t))\\
                 &=(x_0,y_0,z_0)(1-t)\\
                 &+(x_1,y_1,z_1)t, t\in [0,1].
\end{split}
\label{eq:ray-segment}
\end{equation}
We want the interpolation function (basis) to be normalized given fixed $\mathbf{x}_0,\mathbf{x}_1$; the normalized $\mathbf{f}(\mathbf{x}(t))$ along the ray is:
\begin{equation}
    \begin{split}
    \hat{\mathbf{f}}(\mathbf{x}(t)) &=
    \frac{\mathbf{f}(\mathbf{x}(t))}{\int_{\mathbf{x}_0}^{\mathbf{x}_1} \sum_{k=1}^8 \chi_k(\mathbf{x}(t))dt}\\
    &= \frac{\mathbf{f}(\mathbf{x}(t))}{\int_0^1 \sum_{k=1}^8 \chi_k(x(t),y(t),z(t))\lVert \mathbf{x}'(t) \rVert_2 dt}\\
    &= \frac{\mathbf{f}(\mathbf{x}(t))}{\lVert \mathbf{x}_1-\mathbf{x}_0 \rVert_2},
    \end{split}
    \label{eq:trilinear-norm}
\end{equation}
which gives normalized feature integration:
\begin{equation}
\begin{split}
    \int_{\mathbf{x}_0}^{\mathbf{x}_1} \hat{\mathbf{f}}(\mathbf{x}(t))dt
    &=\int_0^1 \frac{\mathbf{f}(x(t),y(t),z(t))}{\lVert \mathbf{x}_1 - \mathbf{x}_0 \rVert_2} \lVert \mathbf{x}'(t) \rVert dt\\
    &= \int_0^1\sum_{k=1}^8 \mathbf{f}_k \chi_k(x(t),y(t),z(t))dt\\
    &= \sum_{k=1}^8 \mathbf{f}_k \int_0^1 \chi_k (x(t),y(t),z(t))dt\\
    &= \sum_{k=1}^8 \mathbf{f}_kX_k(\mathbf{x}_0,\mathbf{x}_1).
\end{split}
\end{equation}
The integration of each interpolation function only depends on the entry and exit that produces a polynomial. In our CUDA implementation, we factorize $X_k(\mathbf{x}_0,\mathbf{x}_1)$ as:
\begin{equation}
    \begin{cases}
        X_8(\mathbf{x}_0,\mathbf{x}_1)&= \frac{2x_0y_0z_0+2x_1y_1z_1+ a b c}{12}\\
        X_7(\mathbf{x}_0,\mathbf{x}_1)&= -X_8(\mathbf{x}_0,\mathbf{x}_1)+  d\\
        X_6(\mathbf{x}_0,\mathbf{x}_1)&= \frac{ a c+x_0z_0+x_1z_1}{6}-X_8(\mathbf{x}_0,\mathbf{x}_1)\\
        X_5(\mathbf{x}_0,\mathbf{x}_1)&= \frac{ c}{2}-X_6(\mathbf{x}_0,\mathbf{x}_1)-  d\\
        X_4(\mathbf{x}_0,\mathbf{x}_1)&= -X_8(\mathbf{x}_0,\mathbf{x}_1)+  e\\
        X_3(\mathbf{x}_0,\mathbf{x}_1)&= \frac{ b}{2}-X_7(\mathbf{x}_0,\mathbf{x}_1)-  e\\
        X_2(\mathbf{x}_0,\mathbf{x}_1)&= \frac{a}{2} - X_6(\mathbf{x}_0,\mathbf{x}_1) -   e\\
        X_1(\mathbf{x}_0,\mathbf{x}_1)&= 1-\frac{ a+b}{2}-X_5(\mathbf{x}_0,\mathbf{x}_1)+  e\\
    \end{cases},
\end{equation}
where:
\begin{equation}
    \begin{cases}
         a&=x_0+x_1\\
         b&=y_0+y_1\\
         c&=z_0+z_1\\
         d&= \frac{ b c+y_0z_0+y_1+z_1}{6}\\
         e&= \frac{ a b+x_0y_0+x_1y_1}{6}
    \end{cases}.
\end{equation}

\subsection{Real-time ray-voxel intersection}
\label{subsec:ray-voxel-intersection}
For each ray, we first find its closest hit on the voxel grid, and then ray marching to a fixed number of voxels in each iteration of MLP evaluation. For closest hit calculation, we build an octree from the occupancy map as a list of 3D arrays by max pooling and traverse in the octree to speed up the calculation. For ray marching after the closest hit, we do not use the octree and use the algorithm described in~\cite{Amanatides1987AFV}.

\subsection{Real-time MLP evaluation} 
\label{subsec:mlp-evaluation}
Because weights and biases of the decoder MLP are globally shared, we upload them to CUDA constant memory to speed up the memory read. Additionally, we refactor two linear layers in the MLP to reduce calculations. We use the \ours32 decoder architecture for the illustration, which can be easily extended to \ours64.

\vspace{-4mm}
\paragraph{Pre-multiplication of the first layer:} 
Because there is no activation (ReLU) between the integrated feature and the first layer of the MLP, the weight of the first layer can be pre-multiplied to the feature vectors. Given the weight and bias of the first linear layer as $\mathbf{W}_1,\mathbf{b}_1$, the first layer's output $\mathbf{e}_1$ (without activation) is:
\begin{gather}
    \begin{split}
        \mathbf{e}_1 &= \mathbf{W}_1 \int_{\mathbf{x}_0}^{\mathbf{x}_1} \hat{\mathbf{f}}(\mathbf{x}(t))dt+\mathbf{b}_1\\
                    &= \mathbf{W}_1\sum_{k=1}^8 \mathbf{f}_k \mathbf{X}_k(\mathbf{x}_0,\mathbf{x}_1)+\mathbf{b}_1\\
            &= \sum_{k=1}^8 \mathbf{f}'_k \mathbf{X}_k(\mathbf{x}_0,\mathbf{x}_1) + \mathbf{b}_1\\
            &= \int_{\mathbf{x}_0}^{\mathbf{x}_1} \hat{\mathbf{f}'}(\mathbf{x}(t))dt + \mathbf{b}_1,
    \end{split}
    \label{eq:pre-multiply}
\end{gather}
where:
\begin{equation}
        \mathbf{f}'_k = \mathbf{W}_1 \mathbf{f}_k.
\end{equation}
By pre-multiplying the weight to each feature vector after the training and using Eq.~\ref{eq:pre-multiply} during inference time, the operation needed for evaluating the first layer is reduced to a vector add.

\vspace{-4mm}
\paragraph{Composition of the third and fourth layers:} 
Similarly, the hidden feature $\mathbf{h}_3$ of the third layer is not mapped with ReLU, such that weights in the third and fourth layers can be composited. Let $\mathbf{W}_3, \mathbf{b}_3$ denote the weight and bias of the third layer, and $\mathbf{W}_4,\mathbf{b}_4$ denote the weight and bias of the fourth layer. Given the hidden feature of the second layer $\mathbf{h}_2$ and the positional encoded viewing direction $\gamma(\mathbf{d})$, we have:
\begin{gather}
    \begin{split}
        \begin{bmatrix}
            \sigma\\
            \mathbf{h}_3
        \end{bmatrix} &= \mathbf{W}_3 \mathbf{h}_2 + \mathbf{b}_3\\
        &=
        \begin{bmatrix}
            \mathbf{W}_3^\sigma \\
            \mathbf{W}_3^\mathbf{h}
        \end{bmatrix}
        \mathbf{h}_2+
        \begin{bmatrix}
            \mathbf{b}_3^\sigma \\
            \mathbf{b}_3^\mathbf{h}
        \end{bmatrix}
    \end{split}\\
    \begin{split}
        \mathbf{e}_4 &= \mathbf{W}_4
        \begin{bmatrix}
            \gamma(\mathbf{d})\\
            \mathbf{h}_3
        \end{bmatrix} + \mathbf{b}_4\\
        &= \begin{bmatrix}
            \mathbf{W}_4^\mathbf{d} & \mathbf{W}_4^\mathbf{h}
        \end{bmatrix}
        \begin{bmatrix}
            \gamma(\mathbf{d})\\
            \mathbf{h}_3
        \end{bmatrix} + \mathbf{b}_4\\
        &= \mathbf{W}_4^\mathbf{d} \gamma(\mathbf{d})
        + \mathbf{W}_4^\mathbf{h} \mathbf{h}_3 + \mathbf{b}_4\\
        &= \mathbf{W}_4^\mathbf{d} \gamma(\mathbf{d})
        + \mathbf{W}_4^\mathbf{h} 
        (\mathbf{W}_3^\mathbf{h} \mathbf{h}_2 + \mathbf{b}_3^\mathbf{h}) + \mathbf{b}_4\\
        &= \mathbf{W}_4^\mathbf{d} \gamma(\mathbf{d})
        + (\mathbf{W}_4^\mathbf{h} \mathbf{W}_3^\mathbf{h})\mathbf{h}_2
        + (\mathbf{W}_4^\mathbf{h}\mathbf{b}^\mathbf{h}_3+\mathbf{b}_4).
    \end{split}
\end{gather}
Therefore, the density $\sigma$ and hidden feature of the fourth layer $\mathbf{e}_4$ (without activation) could be directly calculated from $\gamma(\mathbf{d})$ and $\mathbf{h}_2$ without evaluating $\mathbf{h}_3$:
\begin{gather}
    \label{eq:matrix-composition}
    \sigma = \mathbf{W}_3^\sigma \mathbf{h}_2 + \mathbf{b}^\sigma_3
    \\
    \mathbf{e}_4 = \mathbf{W}_4^\mathbf{d} \gamma(\mathbf{d})
                 + \mathbf{W}_4' \mathbf{h}_2
                 + \mathbf{b}_4'\\
    \begin{split}
        \text{where:}&\\
        \mathbf{W}'_4 =\mathbf{W}_4^\mathbf{h} \mathbf{W}_3^\mathbf{h}
        \text{ and } &
        \mathbf{b}_4' = \mathbf{W}_4^\mathbf{h}\mathbf{b}_3^\mathbf{h} + \mathbf{b}_4,
    \end{split}
\end{gather}
which avoids one $32\times32$ matrix multiplication and one $32$ dimension vector add.

\subsection{Object swapping}
\label{subsec:object-swapping}
We use two cuboids to mark the objects to be swapped and run k-mean clustering for each region to get the fine segmentation. Feature vectors that belong to the largest cluster are treated as the background; the rest of the features are treated as the foreground objects to be swapped. In the hot-dog scene, we use 12 clusters.

%% file: appendix/experiment_details.tex
\section{Experiment Details}
\label{experiment-details}
In Tab.~\ref{tab:detail-synthetic-nerf}, we show the per-scene rendering quality comparison on the NeRF-synthetic dataset for all the baselines we compared with (offline, real-time pre-trained, and real-time applications). Tab.~\ref{tab:detail-tnt} shows the per-scene offline rendering quality on the Tanks and Temple and BlendedMVS datasets, and Tab.~\ref{tab:detail-performance} shows the per-scene real-time performance on the NeRF-synthetic dataset. For ablation on the network architecture, we also show the per-scene performance and rendering quality in Tab.~\ref{tab:detail-alblation}.

\newpage
\input{tables/detail_synthetic_nerf}
\input{tables/detail_tnt}
\input{tables/detail_performance}
\input{tables/detail_ablation}

%% file: tables/detail_synthetic_nerf.tex
\begin{table*}[t]
\centering
\resizebox{0.85\linewidth}{!}{
\begin{tabular}{l c c c c c c c c c}
\toprule
\multicolumn{10}{c}{PSNR $\uparrow$}\\
    & Chair & Drums & Ficus & Hotdog & Lego & Materials & Mic & Ship & \emph{Mean} \\ 
\midrule
 NeRF & 33.00 & 25.01 & 30.13 & 36.18 & 32.54 & 29.62 & 32.91 & 28.65 & 31.00\\
 JaxNeRF & 33.88 & 25.08 & 30.15 & 36.91 & 33.24 & 30.03 & 34.52 & 29.07 & 31.64\\
 AutoInt & 25.60 & 20.78 & 22.47 & 32.33 & 25.09 & 25.90 & 28.10 & 24.15 & 25.55\\
 NSVF & 33.19 & 25.18 & 31.23 & 37.14 & 32.29 & 32.68 & 34.27 & 27.93 & 31.74\\
 NeRF-SH & 3.98 & 25.17 & 30.72 & 36.75 & 32.77 & 29.95 & 34.04 & 29.21 & 31.57\\
 JaxNeRF+ & 35.35 & 25.65 & 32.77 & 37.58 & 35.35 & 30.29 & 36.52 & 30.48 & 33.00\\
 \midrule
 PlenOctrees&34.66 &25.31 &30.79 &36.79 &32.95& 29.76 &33.97 &29.42 &31.71 \\
 SNeRG& 33.24 &24.57 &29.32 &34.33 &33.82 &27.21 &32.60 &27.97& 30.38\\
 FastNeRF& 32.32&23.75&27.79&34.72&32.28&28.89&31.77&27.69&29.97\\ 
 KiloNeRF&-&-&-&-&-&-&-&-&31.00\\ \hline
 \ours64&34.34	&25.39	&31.77	&36.83	&35.52&	29.63&	34.58&	30.50&32.32\\
 \ours32&34.10&	25.40&	32.03&	36.50&	35.27&	29.25&	34.56&	30.17&32.16\\
 \ours32(RT)&34.09&	25.40&32.02&36.35&	35.17&	29.24&34.53&30.14&32.12\\
\bottomrule
\multicolumn{10}{c}{ }\\
\toprule
\multicolumn{10}{c}{SSIM $\uparrow$}\\
&Chair&Drums & Ficus & Hotdog &Lego & Materials & Mic & Ship & \emph{Mean}\\
\midrule
NeRF &  0.967 &0.925& 0.964 &0.974& 0.961 &0.949& 0.980& 0.856 &0.947\\
JaxNeRF &0.974 &0.927& 0.967& 0.979& 0.968& 0.952& 0.987& 0.865&0.952\\
AutoInt& 0.928& 0.861& 0.898& 0.974& 0.900& 0.930& 0.948 &0.852&0.911\\
NSVF& 0.968 &0.931 &0.973& 0.980& 0.960& 0.973& 0.987& 0.854&0.953\\
NeRF-SH&0.974 &0.927 &0.968 &0.978& 0.966& 0.951& 0.985& 0.866& 0.952\\
JaxNeRF+& 0.982 &0.936& 0.980& 0.983& 0.979& 0.956 &0.991& 0.887&0.962\\ 
\midrule
PlenOctrees&  0.981 &0.933& 0.970& 0.982& 0.971& 0.955& 0.987& 0.884 &0.958\\
SNeRG& 0.975& 0.929 &0.967 &0.971& 0.973& 0.938& 0.982& 0.865& 0.950\\
FastNeRF& 0.966 & 0.913 & 0.954 & 0.973 & 0.964 & 0.947 & 0.977 & 0.805 & 0.941\\
KiloNeRF&-&-&-&-&-&-&-&-&0.950\\ 
\midrule
\ours64& 0.978&	0.933&	0.975&	0.981&	0.980&	0.951&	0.987&	0.893&0.960\\
\ours32& 0.977&	0.932&	0.977&	0.979&	0.979	&0.946&	0.987&0.886	&0.958\\
\ours32(RT)&0.977&	0.932&	0.977&0.978&	0.978&	0.946&0.987&0.885&0.958\\
\bottomrule
\multicolumn{10}{c}{ }\\
\toprule
\multicolumn{10}{c}{LPIPS $\downarrow$}\\
&Chair&Drums & Ficus & Hotdog &Lego & Materials & Mic & Ship & \emph{Mean}\\
\midrule
NeRF &0.046 &0.091 &0.044& 0.121& 0.050& 0.063& 0.028& 0.206 &0.081\\
JaxNeRF&  0.027 &0.070 &0.033 &0.030& 0.030& 0.048& 0.013& 0.156& 0.051\\
AutoInt& 0.141& 0.224& 0.148 &0.080 &0.175& 0.136& 0.131 &0.323& 0.170\\
NSVF&  0.043 &0.069& 0.017 &0.025& 0.029 &0.021& 0.010& 0.162& 0.047\\
NeRF-SH&  0.037& 0.087& 0.039 &0.041& 0.041 &0.060& 0.021& 0.177& 0.063\\
JaxNeRF+& 0.017& 0.057& 0.018 &0.022& 0.017 &0.041 &0.008& 0.123& 0.038\\
\midrule
PlenOctree & 0.022& 0.076 &0.038& 0.032& 0.034& 0.059 &0.017& 0.144& 0.053\\
SNeRG&0.025& 0.061 &0.028& 0.043& 0.022& 0.052& 0.016& 0.156&0.050\\
FastNeRF& 0.032& 0.083 & 0.031& 0.031&0.022&0.034&0.022&0.192&0.053\\
KiloNeRF&-&-&-&-&-&-&-&-&0.030\\
\midrule
\ours64& 0.014&	0.057&	0.020&	0.017&	0.010&	0.032&	0.010&	0.093&	0.032\\
\ours32&0.014&	0.058&	0.020&	0.019&	0.010&	0.035&	0.011&	0.102&	0.034\\
\ours32(RT)&0.014&	0.058&	0.020&	0.019&	0.010&0.034&0.011&	0.100&	0.033\\
\bottomrule
\end{tabular}
}
\caption{\textbf{Rendering quality on the NeRF-synthetic dataset}.}
\label{tab:detail-synthetic-nerf}
\end{table*}

%% file: tables/detail_tnt.tex
\begin{table*}[t]
\centering
\setlength\tabcolsep{3 pt}
\resizebox{0.85\linewidth}{!}{
\begin{tabular}{l cccccc|ccccc}
\toprule
\multicolumn{12}{c}{PSNR $\uparrow$}\\
&Barn & Caterpillar & Family & Ignatius & Truck & \emph{Mean}&
Jade&Fountain&Char&Statues&\emph{Mean}\\
\midrule
NeRF& 24.05	&23.75&	30.29&	25.43&	25.36&	25.78&
21.65&	25.59&	25.87&	23.48&	24.15\\
JaxNeRF& 27.39&	25.24&	32.47&	27.95&	26.66&	27.94&
 -&-&-&-&-\\
NSVF&27.16&	26.44&	33.58&	27.91&	26.92&	28.40&
26.96&	27.73&	27.95&	24.97&	26.90\\
\midrule
\ours64& 27.31&	25.64&	33.40&	27.80&	26.74&	28.18&
26.52&	28.30&	28.81&	25.36&	27.25\\
\bottomrule
\multicolumn{12}{c}{ }\\
\toprule
\multicolumn{12}{c}{SSIM $\uparrow$}\\
&Barn & Caterpillar & Family & Ignatius & Truck & \emph{Mean}&
Jade&Fountain&Char&Statues&\emph{Mean}\\
\midrule
NeRF& 0.750	&0.860&	0.932&	0.920&	0.860&	0.864&
0.750&	0.860&	0.900&	0.800	&0.828\\
JaxNeRF& 0.842&	0.892&	0.951&	0.940&	0.896&	0.904&
 -&-&-&-&-\\
NSVF&0.832	&0.900	&0.954&	0.930&	0.895&	0.900&
0.901&	0.913&	0.921&	0.858&	0.898\\
\midrule
\ours64& 0.850&	0.903&	0.960&	0.941&	0.904&	0.912&
0.900&	0.918&	0.948&	0.873&	0.910\\
\bottomrule
\multicolumn{12}{c}{ }\\
\toprule
\multicolumn{12}{c}{LPIPS $\downarrow$}\\
&Barn & Caterpillar & Family & Ignatius & Truck & \emph{Mean}&
Jade&Fountain&Char&Statues&\emph{Mean}\\
\midrule
NeRF& 0.395&	0.196&	0.098&	0.111&	0.192&	0.198&
0.264&	0.149&	0.149&	0.206&	0.192\\
JaxNeRF& 0.286&	0.189&	0.092&	0.102&	0.173&	0.168&
 -&-&-&-&-\\
NSVF&0.307&	0.141&	0.063&	0.106&	0.148&	0.153&
0.094&	0.113&	0.074&	0.171&	0.113\\
\midrule
\ours64& 0.209&	0.121&	0.050&	0.082&	0.119&	0.116&
0.076&	0.069&	0.037&	0.110&	0.073\\
\bottomrule
\end{tabular}
}
\caption{\textbf{Rendering quality on the Tanks \& Temple and BlendedMVS datasets}.}
\label{tab:detail-tnt}
\end{table*}

%% file: tables/detail_performance.tex
\begin{table*}[t]
\centering
\setlength\tabcolsep{3 pt}
\resizebox{0.75\linewidth}{!}{
\begin{tabular}{l c c c c c c c c c}
\toprule
\multicolumn{10}{c}{FPS $\uparrow$}\\
&Chair&Drums & Ficus & Hotdog &Lego & Materials & Mic & Ship & \emph{Range}\\
\midrule
PlenOctrees& 143&	78&	23&	15&	45&	13&	76	&10 & 76$\pm$66\\
SNeRG& -&-&-&-&-&-&-&-&98$\pm$37\\
FastNeRF&-&-&-&-&-&-&-&-&-\\
KiloNeRF&40&-&-&-&40&-&-&16&28$\pm$12\\
\ours32(RT)&59&	40	&39&	44&	67&	29&	66&	27&47$\pm$20\\
\bottomrule
\multicolumn{10}{c}{ }\\
\toprule
\multicolumn{10}{c}{MB $\downarrow$}\\
&Chair&Drums & Ficus & Hotdog &Lego & Materials & Mic & Ship & \emph{Mean}\\
\midrule
PlenOctrees& 832&	1239&	1792&	2683&	2068&	3686&	443&	2693&	1930\\
SNeRG& -&-&-&-&-&-&-&-&84\\
FastNeRF&-&-&-&-&-&-&-&-&-\\
KiloNeRF&204&-&-&-&108&-&-&173&161\\
\ours32(RT)&55&	56&	47&	84	&64&	62&	24&	151	&68\\
\bottomrule
\multicolumn{10}{c}{ }\\
\toprule
\multicolumn{10}{c}{GPU GB $\downarrow$}\\
&Chair&Drums & Ficus & Hotdog &Lego & Materials & Mic & Ship & \emph{Range}\\
\midrule
PlenOctrees& 0.94&	1.34	&1.87&	2.73&	2.19&	3.70&	0.56&	2.74&	1.65$\pm$1.09\\
SNeRG& -&-&-&-&-&-&-&-& 1.73$\pm$1.48\\
FastNeRF&-&-&-&-&-&-&-&-&-\\
KiloNeRF&1.94&-&-&-&1.41&-&-&1.78&1.68$\pm$0.27\\
\ours32(RT)&1.04&	1.04&	1.03&	1.06&	1.04&	1.04&	1.01&	1.13&1.07$\pm$0.06\\
\bottomrule
\end{tabular}
}
\caption{\textbf{Performance of real-time applications on the NeRF-synthetic dataset}.}
\label{tab:detail-performance}
\end{table*}

%% file: tables/detail_ablation.tex
\begin{table*}[t]
\centering
\setlength\tabcolsep{3.5 pt}
\resizebox{0.85\linewidth}{!}{
\begin{tabular}{ll c c c c c c c c c}
\toprule
\multicolumn{11}{c}{PSNR $\uparrow$}\\
$N$ & Decoder &Chair&Drums & Ficus & Hotdog &Lego & Materials & Mic & Ship & \emph{Mean}\\
\midrule
256 & \ours64(RT) & 34.35	&25.38&	31.76&	36.76&	35.49&	29.61&	34.57&	30.48&	32.30\\
256 & \ours32(RT) &34.09&	25.40&	32.02&	36.35&	35.17&	29.24&	34.53&	30.14&	32.12\\
128 & \ours64(RT) &31.98&	24.74&	30.12&	35.54&	32.57&	28.96&	32.15&	29.02&	30.63\\
128 & \ours32(RT) &31.54&	24.75&	30.25&	35.42&	32.61&	28.82&	31.97&	28.80&	30.52\\
\bottomrule
\multicolumn{11}{c}{ }\\
\toprule
\multicolumn{11}{c}{FPS $\uparrow$}\\
$N$ & Decoder &Chair&Drums & Ficus & Hotdog &Lego & Materials & Mic & Ship & \emph{Range}\\
\midrule
256 & \ours64(RT) & 31	&25	&18&	19&	28&	16&	35&	17& 26$\pm$9\\
256 & \ours32(RT) & 59	&40	&39&	44&	67&	29&	66&	27& 47$\pm$20\\
128 & \ours64(RT) & 57	&38& 29&	33&	41&	28&	53&	17& 37$\pm$20\\
128 & \ours32(RT) & 108&	82&	61&	84&	99&	67&	119&	45& 82$\pm$37\\
\bottomrule
\multicolumn{11}{c}{ }\\
\toprule
\multicolumn{11}{c}{MB $\downarrow$}\\
$N$ & Decoder &Chair&Drums & Ficus & Hotdog &Lego & Materials & Mic & Ship & \emph{Mean}\\
\midrule
256 & \ours64(RT) & 55	&42	&49	&80&	64&	62&	24&	118	&62\\
256 & \ours32(RT) & 55	&56&	47	&84&	64&	62&	24&	151	&68\\
128 & \ours64(RT) & 9.2	&8.2	&8.5	&15&	12&	9.6&	4.7	&30&	12\\
128 & \ours32(RT) &9.7&	8.9&	9.3	&16&	13&	9.8&	4.5&	28&	12\\
\bottomrule
\end{tabular}
}
\caption{\textbf{Architecture ablation on the NeRF-synthetic dataset}.}
\label{tab:detail-alblation}
\end{table*}